%% file: main.tex
\RequirePackage{fix-cm}
\PassOptionsToPackage{sort&compress}{natbib}
\PassOptionsToPackage{table,dvipsnames}{xcolor}
\documentclass[11pt,a4paper]{jitagent}
\input{preamble}

\title{\raisebox{-0.15em}{\includegraphics[height=0.95em]{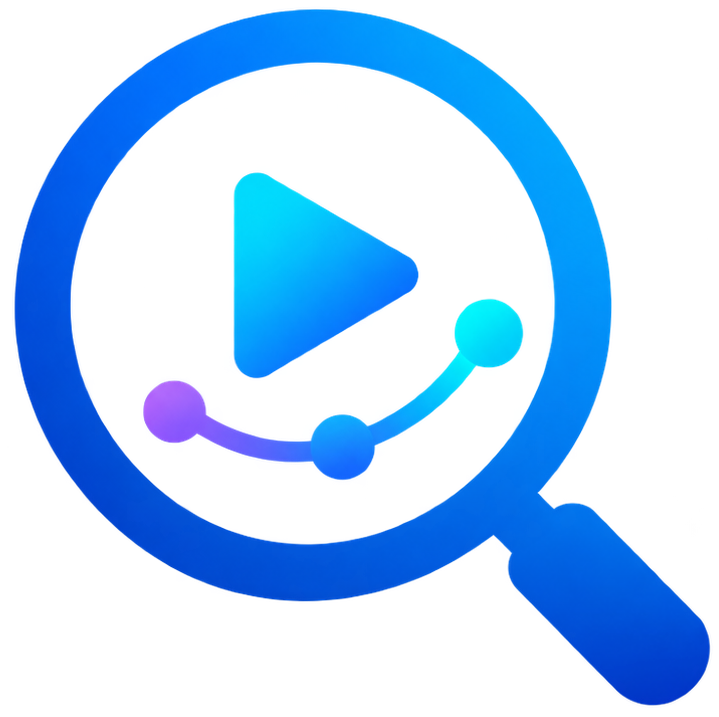}}\hspace{0.2em}Loupe: Reinforcing Knowledge-Intensive\\ Video Generation Agents}
\author{%
  Kai Zeng\amark{1,2,*}, Ruichuan An\amark{2,*}, Zhen Fang\amark{3}, Xi Fang\amark{1,\dagger}, Han Lyu\amark{1}, Moran Li\amark{4}\\
  Yingchen Yu\amark{5}, Yifan Dai\amark{6}, Bohan Zeng\amark{2}, Quanqing Xu\amark{7}, Zhengzhou Zhu\amark{2}, Wentao Zhang\amark{2,\ddagger}\\[6pt]
  {\normalfont\small \amark{1}DP Technology\quad \amark{2}PKU\quad \amark{3}USTC\quad \amark{4}HUST\quad \amark{5}NTU\quad \amark{6}SJTU\quad \amark{7}Ant Group}\\[3pt]
  {\normalfont\footnotesize \amark{*}Equal contribution\quad \amark{\dagger}Project leader\quad \amark{\ddagger}Corresponding author}}
\date{}
\metadata[\textcolor{atriablue}{\faEnvelope}\ Contact]{\href{mailto:asbeforekz@gmail.com}{\textcolor{black}{asbeforekz@gmail.com}}, \href{mailto:wentao.zhang@pku.edu.cn}{\textcolor{black}{wentao.zhang@pku.edu.cn}}}
\metadata[\textcolor{atria}{\faGlobe}\ Project Page]{\url{https://zengkaiya.github.io/Loupe/}}
\hypersetup{
  pdftitle={Loupe: Reinforcing Knowledge-Intensive Video Generation Agents},
  pdfauthor={Kai Zeng, Ruichuan An, Zhen Fang, Xi Fang, Han Lyu, Moran Li, Yingchen Yu, Yifan Dai, Bohan Zeng, Quanqing Xu, Zhengzhou Zhu, Wentao Zhang}}
\begin{document}
\maketitle
\input{sec/0_abstract}
\input{sec/1_intro}
\input{sec/2_related_work}
\input{sec/3_method}
\input{sec/4_results}
\input{sec/5_conclusions}


\clearpage
\pdfbookmark[1]{References}{references}
{\small
\bibliographystyle{ieeenat_fullname}
\bibliography{main}
}
\clearpage
\appendix
\crefalias{section}{appendix}
\crefalias{subsection}{appendix}
\input{sec/X_suppl}

\end{document}

%% file: preamble.tex
\usepackage{atria-jit-branding}
\usepackage{fontawesome5}
\usepackage{makecell,multirow,placeins,needspace}
\usepackage[capitalize,nameinlink]{cleveref}
\crefformat{table}{#2Tab.#1#3}
\Crefformat{table}{#2Tab.#1#3}
\crefformat{figure}{#2Fig.#1#3}
\Crefformat{figure}{#2Fig.#1#3}
\crefname{appendix}{Appx.}{Appx.}
\Crefname{appendix}{Appx.}{Appx.}
\crefformat{appendix}{#2Appx.#1#3}
\Crefformat{appendix}{#2Appx.#1#3}
\setcitestyle{numbers,square,comma}
\definecolor{atria}{HTML}{2456A6}
\definecolor{atriablue}{HTML}{2F5E9E}
\definecolor{atrialight}{HTML}{E9EEF7}
\hypersetup{colorlinks=true,linkcolor=atria,citecolor=atria,urlcolor=atria}
\newcommand{\amark}[1]{\textsuperscript{\normalfont\ensuremath{#1}}}
\pretitle{\par\begingroup\centering\fontsize{17pt}{20pt}\selectfont\bfseries}
\posttitle{\par\endgroup\noindent{\color{black}\rule{\textwidth}{0.4pt}}\par}
\preauthor{\par\begingroup\centering\bfseries\fontsize{9.9pt}{13pt}\selectfont}
\postauthor{\par\endgroup}
\renewcommand{\jitabstractfont}{\fontsize{9.2pt}{11.7pt}\selectfont}

\definecolor{red4}{HTML}{febf92}
\definecolor{blue4}{HTML}{77b1e2}
\definecolor{green4}{HTML}{F5FFFA}
\definecolor{reqbrown}{HTML}{D18453}
\definecolor{reqyellow}{HTML}{E5AD00}
\definecolor{reqcyan}{HTML}{35B3AD}
\definecolor{reqblue}{HTML}{527DE1}
\definecolor{promptblue}{HTML}{EAF3FC}
\definecolor{promptborder}{HTML}{96BCE2}
\definecolor{gainGreen}{RGB}{0,176,0}
\definecolor{taxonomyHeader}{HTML}{E8EEFC}
\definecolor{taxonomyStripe}{HTML}{F6F7F8}
\newtcolorbox{mybreakablepromptbox}[1][]{
  colback=promptblue, colframe=promptborder,
  colbacktitle=promptborder!35!white, coltitle=black,
  boxrule=0.5pt, arc=2mm,
  left=6pt, right=6pt, top=5pt, bottom=5pt,
  fontupper=\small\ttfamily, fonttitle=\small\bfseries,
  title={#1}, breakable, enhanced,
  before upper={\raggedright}}
\DeclareUnicodeCharacter{300A}{\ensuremath{\langle\!\langle}}
\DeclareUnicodeCharacter{300B}{\ensuremath{\rangle\!\rangle}}

\newcommand{\gain}[1]{\,{\color{gainGreen}\scriptsize$\uparrow$#1}}
\newcommand{\scidrop}[1]{\,{\color{dropRed}\scriptsize$\downarrow$#1}}
\definecolor{dropRed}{RGB}{220,50,50}

%% file: sec/0_abstract.tex
\begin{abstract}
Knowledge-intensive video generation remains challenging because prompts often omit the facts, appearances, and temporal relationships required for accurate generation. 
While search-based workflows can retrieve missing evidence, manually designed strategies limit adaptive planning, and conventional training suffers from costly video-level feedback and coarse trajectory-level credit assignment. 
In this paper, we present \textbf{Loupe}, a video generation agent that retrieves and verifies textual and visual evidence to construct evidence-grounded temporal video prompts with reference images. 
We further introduce \textbf{Rubric-Based Credit (RBC)}, which leverages prompt-specific factual, visual, and temporal requirements to evaluate intermediate search and planning decisions before video rendering. 
Building on RBC, \textbf{RBC-SFT} reweights assistant-token learning by step quality, while \textbf{RBC-GRPO} incorporates step-level credits alongside a final planning reward, without rendering videos during RL. 
To address the lack of dedicated search-oriented evaluation, we construct \textbf{Search-VBench}, which contains 150 held-out prompts across seven categories.
Experiments show that Loupe improves all four evaluation dimensions across three video generators, raising the overall score of MiniMax-H3 from 32.80 to 70.67 and outperforming an untrained search workflow.
Ablation experiments show that full RBC training outperforms standard SFT followed by trajectory-level GRPO by 5.27 points.
Human evaluations and results on Sci-VBench and KIVI-Bench further demonstrate improved generation quality, scientific grounding, factual precision, and helpfulness.
\printprojectmetadata
\end{abstract}

\begin{figure}[H]
\centering
\includegraphics[width=0.95\linewidth]{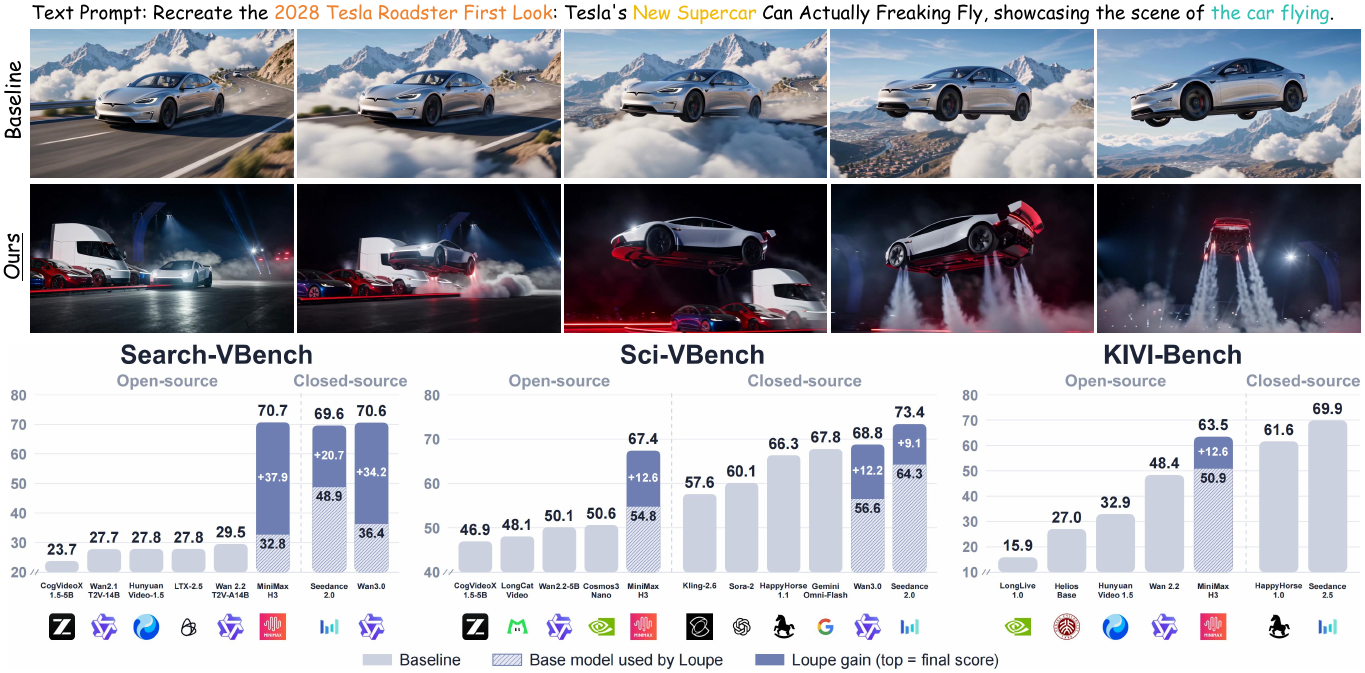}
\caption{\textbf{Loupe helps multiple backbones address knowledge-intensive video generation.}
\textbf{Top}: direct generation is prone to hallucinated content because of \textcolor{reqbrown}{know}\textcolor{reqyellow}{ledge} \textcolor{reqcyan}{gaps}, whereas Loupe's multi-hop search and planning acquire specific entity appearances and long-tail factual knowledge.
\textbf{Bottom}: results on Search-VBench, Sci-VBench, and KIVI-Bench consistently show the gains from Loupe. Owing to space limitations, we show only a subset of the results; the complete results are provided in the experiments section.
}
\label{fig:teaser}
\end{figure}

%% file: sec/1_intro.tex
\section{Introduction}
\label{sec:intro}

Video generation models~\cite{minimaxh3,seedance2026seedance,wan30} have advanced rapidly in visual fidelity and motion quality. However, knowledge-intensive video generation~\cite{zhang2026sci,wang2026knowledge} remains challenging because requests often omit the factual details, distinctive appearances, and temporal relationships needed to depict a named entity, event, or process. Consequently, generators may produce visually compelling videos while hallucinating entities, misrepresenting appearances, or depicting incorrect event sequences (\cref{fig:teaser}). Improving rendering capabilities alone cannot address these knowledge gaps.

Equipping video generation agents~\cite{huang2026vimax,feng2026newton,li2026videogen} with search tools offers a promising solution. However, effective evidence acquisition requires agents to identify knowledge gaps, formulate queries, verify information, select visual references, and integrate evidence into coherent generation plans. Existing prompt-based workflows rely largely on manually designed strategies, limiting adaptive search and evidence reasoning. This motivates learning agents that can make effective search and planning decisions.

However, training such agents introduces a credit assignment challenge. Conventional supervised fine-tuning (SFT) treats assistant tokens uniformly despite varying decision quality, while reinforcement learning (RL) based on final-video feedback is costly and entangles agent decisions with the generator's rendering ability. Moreover, trajectory-level rewards obscure individual step contributions. Recent methods~\cite{lu2026abseeker,xie2026step} explore process supervision through answer-backtracked clues or rubrics derived from correctness-contrasted reasoning trajectories. However, knowledge-intensive video requests typically lack annotated factual answers or target videos, making such supervision particularly difficult to obtain directly in practice.

Our key insight is that \textbf{a request specifies what must be verified, even when the corresponding answers are unknown}. It may require identifying an entity, establishing factual details, grounding its appearance, and preserving event order without providing the answers. These requirements define evaluation criteria rather than factual answers, which must be acquired through external evidence. Combining request-derived criteria with interaction history enables us to evaluate intermediate decisions \textbf{without requiring annotated factual answers or target videos}.

Based on this insight, we present \textbf{Loupe}, a video generation agent that retrieves and integrates external evidence into grounded generation inputs (\cref{fig:overview}). Loupe identifies knowledge gaps and iteratively uses text search, image search, and web browsing to acquire and verify evidence, producing temporally coherent video prompts and supporting reference images for downstream generators.

To train Loupe, we introduce \textbf{Rubric-Based Credit (RBC)}, which evaluates intermediate decisions against request-derived requirements and retrieved evidence. For each prompt, RBC constructs a rubric comprising \emph{reference requirements}, specifying the entities, facts, visual attributes, and temporal relations to establish, and \emph{grounding types}, characterizing associated knowledge needs. A process judge evaluates each assistant step using the rubric and interaction history, assigning credit according to its contribution to evidence acquisition and utilization. Building on RBC, \textbf{RBC-SFT} reweights assistant-token loss according to normalized step credits, while \textbf{RBC-GRPO} incorporates step-specific credits into relative advantage estimation across corresponding turn positions, alongside a final reward for the completed video plan and reference images. Both rewards are evaluated without video rendering.

To evaluate knowledge-intensive video generation, we construct \textbf{Search-VBench}, a dedicated benchmark comprising 150 held-out prompts across seven categories, 339 reference images, and 566 evaluation rubric items. It measures faithfulness, visual correctness, temporal dynamics, and aesthetics. Experiments demonstrate that Loupe improves all four dimensions across three video generators, raising MiniMax-H3's~\cite{minimaxh3} overall score from 32.80 to 70.67 and outperforming an untrained search workflow. Human evaluations yield an overall preference rate of 82.33, while evaluations on \textbf{Sci-VBench}~\cite{zhang2026sci} and \textbf{KIVI-Bench}~\cite{wang2026knowledge} demonstrate gains in scientific grounding, factual precision, and helpfulness. Moreover, ablation experiments show that full RBC training outperforms standard SFT followed by trajectory-level GRPO by 5.27 points.

Our contributions are summarized as follows:
\begin{itemize}
    \item We introduce \textbf{Loupe}, an evidence-driven video generation agent that retrieves and verifies external textual and visual evidence to produce grounded temporal prompts and reference images across multiple generators.

    \item We propose \textbf{Rubric-Based Credit (RBC)}, a request-conditioned process supervision framework that derives evaluation criteria without annotated factual answers. Together with RBC-SFT and RBC-GRPO, it enables step-level supervision in supervised and reinforcement learning without video rendering during training.

    \item We construct \textbf{Search-VBench}, a dedicated benchmark for knowledge-intensive video generation, and demonstrate Loupe's effectiveness across multiple video generators, benchmarks, and human evaluations.
\end{itemize}

%% file: sec/2_related_work.tex
\section{Related Work}
\label{sec:related_work}

\noindent \textbf{Agentic image generation.}
Early systems such as GenArtist~\cite{wang2024genartist} and GEMS~\cite{he2026gems} rely on prompt-based workflows.
Gen-Searcher~\cite{feng2026gen} learns search-augmented generation with textual and image feedback.
Loupe extends evidence acquisition to video, where retrieved knowledge must support actions, state changes, and temporal relationships, and supervises search and planning before rendering.

\noindent \textbf{Agentic video generation.}
Existing systems explore task decomposition, staged planning, and external tools~\cite{huang2026vimax,wei2026videoweaver,feng2026newton,li2026videococo}.
AgenticGen~\cite{bu2026agenticgen} optimizes advertising strategy selection and draft generation using business performance.
VideoGen-Agent~\cite{li2026videogen} learns tool coordination through SFT and RL, with rewards incorporating generated-video quality.
In contrast, Loupe focuses on external evidence acquisition and temporal planning, using step-level credits and a final planning reward without rendering videos during RL.

\noindent \textbf{Process supervision and credit assignment.}
Prior work explores information gain, critical-step verification, and intermediate evidence structures for search agents~\cite{wang2026information,li2026verified,liu2026beyond,team2026mind}. ABSeeker~\cite{lu2026abseeker} backtracks from known correct answers to construct supervision clues for search steps, while SRaR~\cite{xie2026step} derives rubrics from answer-verified correct and incorrect reasoning trajectories and combines process feedback with answer-based rewards. 
Unlike ABSeeker and SRaR, whose original supervision construction relies on known answers or answer-verified trajectories, Loupe derives verification requirements directly from user requests and evaluates individual agent decisions against these requirements and retrieved evidence, enabling process supervision without annotated factual answers or target videos.

%% file: sec/3_method.tex
\section{Loupe}
\label{sec:method}
In Sec.~\ref{sec:agent}, we describe the inference process of Loupe.
In Sec.~\ref{sec:rbc}, we elaborate on how Rubric-Based Credit is obtained (\cref{fig:score_details}).
In Secs.~\ref{sec:sft} and~\ref{sec:grpo}, we present how RBC is used for fine-grained SFT and RL.
Further details of the training data construction process are provided in \cref{sec:supp_data}.

\begin{figure}[ht]
\centering
\includegraphics[width=0.95\linewidth]{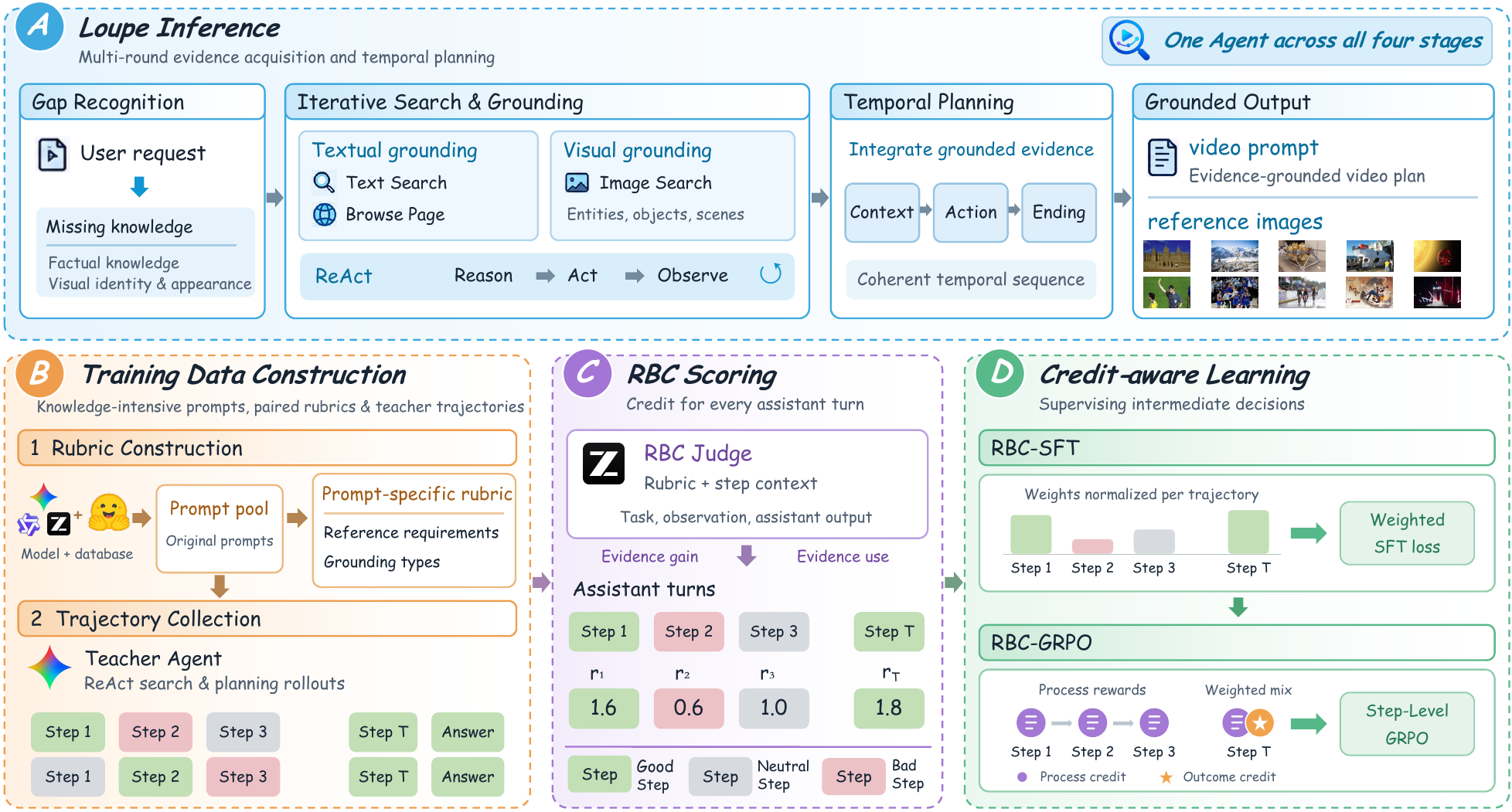}
\caption{\textbf{Overview of Loupe.}
(A) The agent retrieves evidence and produces a temporal video prompt with reference images.
(B) Prompt, rubric, and teacher trajectory construction enables (C) Rubric-Based Credit (RBC) assignment and (D) RBC-SFT and RBC-GRPO training.}
\label{fig:overview}
\end{figure}

\subsection{Search and Temporal Planning}
\label{sec:agent}
An underspecified request requires the agent to determine which facts, appearances, and actions need external grounding. 
As shown at the top of \cref{fig:overview}, given a video request $x$, Loupe searches for missing evidence and returns a temporal video prompt $p$ together with selected reference images $I$. A downstream generator renders the video from $(p,I)$. 
Loupe follows a ReAct interaction loop~\cite{yao2022react} with three tools. \textit{Text Search} retrieves facts, entity relationships, scientific mechanisms, and procedural steps. \textit{Image Search} supplies visual references for entities and scenes, while \textit{Browse Page} inspects retrieved pages to verify details or resolve ambiguity. 
The same agent chooses queries, interprets tool observations, and revises its plan across turns.

\Needspace{2\baselineskip}
Let $a_t$ denote the assistant output at turn $t$, and let $h_t$ contain the request and the preceding interaction, including available tool observations. A trajectory contains $T$ assistant turns; its final output $a_T=(p,I)$ describes the scene and its temporal order. Planning must translate retrieved facts into the depicted actions and state changes, rather than simply append search results to the original request. 

\subsection{Rubric-Based Credit}
\label{sec:rbc}
Evaluating only the final plan or the video leaves the contribution of earlier search steps unresolved.
We observe that a video can be realized in multiple correct ways, yet it must satisfy a set of verifiable factual, appearance, and temporal requirements. 
These requirements can serve as the basis for supervising the search process while reducing interference from the video generator's rendering capability during agent training.
For each request, we therefore construct a rubric $q$ with two fields: \textit{reference requirements} and \textit{grounding type}.
The requirements identify the entities, facts, visual attributes, and temporal relationships that the video must satisfy.
The grounding type describes the associated knowledge need, such as entity discovery, scientific mechanism, or procedural order.
Note that none of the rubrics contains predefined answers; they only specify the aspects to check when assessing search progress.

\begin{figure}[t]
\centering
\includegraphics[width=0.77\linewidth]{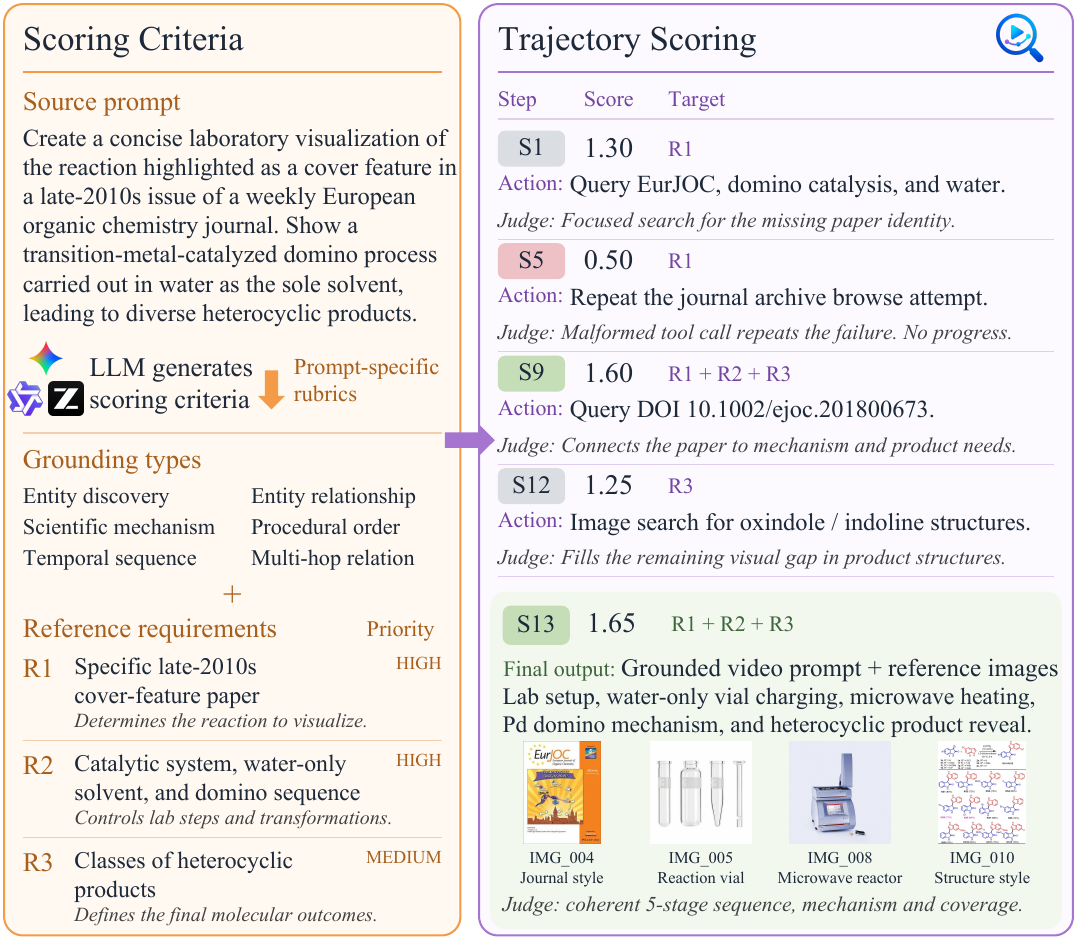}
\caption{\textbf{An illustrative example of RBC scoring.} \textbf{Left}: the request is decomposed into multiple grounding types and reference requirements. \textbf{Right}: selected turns show how the judge scores the search process and the final plan based on prompt-specific rubrics.}
\label{fig:score_details}
\end{figure}

A judge evaluates $(x,q,h_t,a_t)$ to determine whether turn $t$ advances an unresolved requirement, correctly uses evidence, repeats an unproductive action, or introduces unsupported information. Note that the tool observations from the current turn are not provided to the judge; they are used to assess evidence use in the next turn.
Credit starts at the neutral value of one and is adjusted by the rubric items:
\begin{equation}
r_t=\operatorname{clip}\!\left(1+\sum_{j\in\mathcal A_t}\Delta_j,\,0,\,2\right),
 \label{eq:rbc}
\end{equation}
where $\mathcal A_t$ is the set of applicable scoring items and $\Delta_j$ is an item's adjustment.
The result is one score per assistant turn, including the final plan.
The concrete adjustments are listed in \cref{tab:rubric_scoring}.
The judge rewards focused information seeking, verified factual or visual evidence, and correct synthesis.
It penalizes irrelevant tool use, contradictions, and omitted requirements.
A search call is not itself evidence acquisition, but a reasonable query is not penalized solely because the tool returns no useful result.
For the final plan, fluent wording cannot compensate for unsupported facts or missing requirements.
In the worked example in \cref{fig:score_details}, the rubric separates paper identity, reaction mechanism, and product appearance.
A repeated archive-browsing failure receives 0.50 credit, while retrieving the relevant reaction page receives 1.60.
The final plan receives 1.65 after integrating the required mechanism and visual references.

\begin{table}[htbp]
\centering
\small
\caption{\textbf{RBC adjustments} are added to the neutral credit of 1.}
\label{tab:rubric_scoring}
\setlength{\tabcolsep}{3pt}
\begin{tabular}{p{0.69\linewidth}c}
\toprule
Behavior & $\Delta$ \\
\midrule
Focused text search or reasoning & $[0.2,0.4]$ \\
Targeted image search & $[0.2,0.4]$ \\
Browsing verifies facts or resolves ambiguity & $[0.2,0.4]$ \\
Correct evidence synthesis & $[0.3,0.6]$ \\
Complete, grounded final answer & $[0.3,0.7]$ \\
Vague or irrelevant tool use & $[-0.5,-0.2]$ \\
Unsupported facts or contradictions & $[-0.9,-0.4]$ \\
Missing requirements or invalid references & $[-0.9,-0.4]$ \\
Harmful, deceptive, or unusable output & $\leq-1.0$ \\
\bottomrule
\end{tabular}
\end{table}

\subsection{RBC-SFT: Credit-Weighted Learning}
\label{sec:sft}
\begin{figure}[t]
\centering
\includegraphics[width=0.80\linewidth]{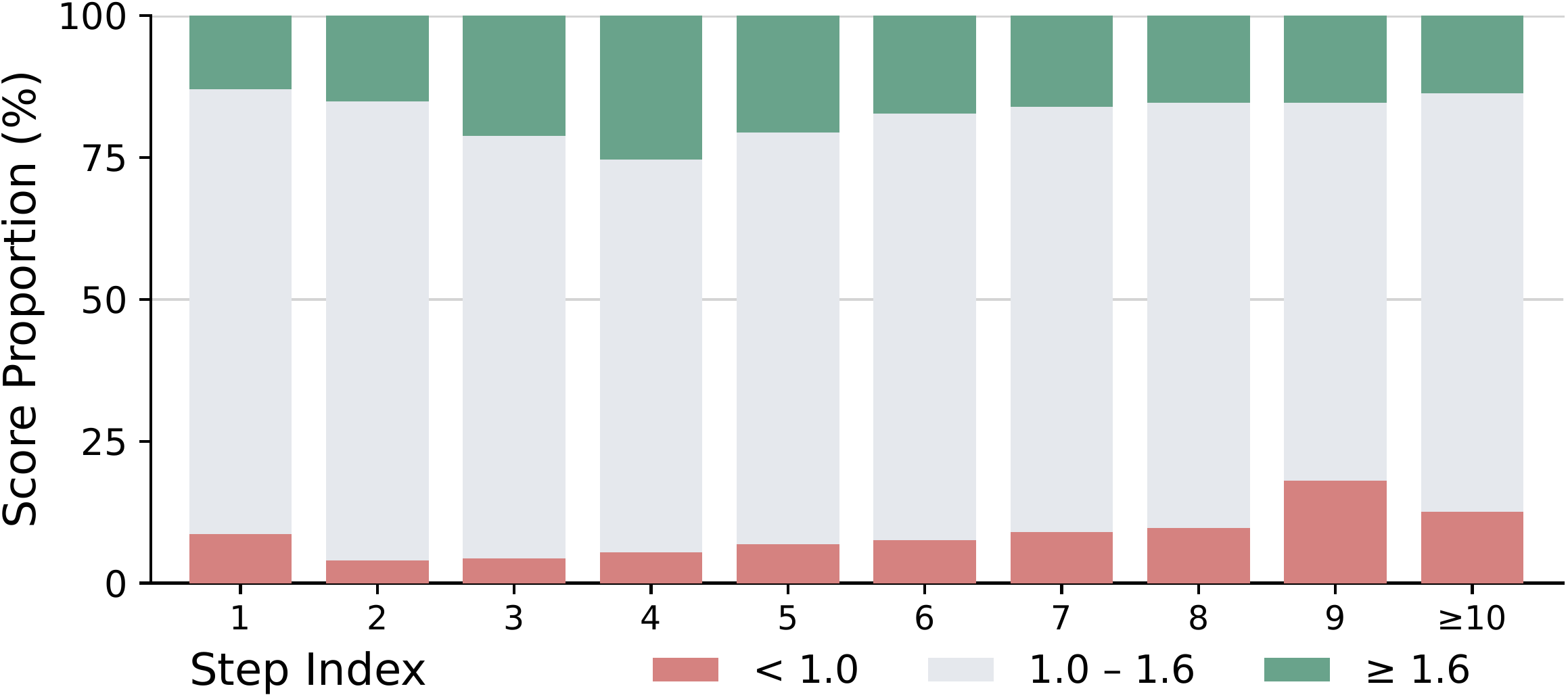}
\caption{\textbf{Distribution of RBC scores across SFT trajectories.} We find that decisions of differing quality can occur at every step, which motivates the use of discriminative supervision signals.}
\label{fig:sft_trajectories}
\end{figure}

As shown in \cref{fig:sft_trajectories}, 
the teacher trajectories contain both useful decisions and local mistakes.
The credit distribution includes low- and high-value turns at every position, so uniform imitation would conceal this within-trajectory variation.
Discarding an entire trajectory instead wastes the useful evidence-seeking behavior it contains.

To address these issues, RBC-SFT retains the trajectory and redistributes its supervision across turns.
We normalize the step credits within each trajectory and use them to compute a weighted assistant-token loss:
\begin{equation}
\begin{gathered}
\mathcal L_{\mathrm{SFT}}(\theta)
=-
\frac{
\sum_{t=1}^{T}w_t\sum_{j=1}^{n_t}
\log\pi_\theta(a_{t,j}\mid h_t,a_{t,<j})
}{
\sum_{t=1}^{T}n_tw_t
},\\[6pt]
w_t=\frac{r_t}{\sum_{s=1}^{T}r_s}.
\end{gathered}
\label{eq:sft}
\end{equation}
Here, $n_t$ denotes the number of supervised assistant tokens in turn $t$.
Tool observations remain context and are excluded from both supervision and normalization.
All tokens in an assistant turn share the same weight, giving higher-credit turns greater influence per token.
We compute the weighted token mean within each trajectory and then average across trajectories in the batch.
The objective reduces to the ordinary SFT loss when all credits are equal and positive.

\subsection{RBC-GRPO: Step-Level Credit Optimization}
\label{sec:grpo}
Starting from RBC-SFT, we sample a group of rollouts for each prompt and score their assistant turns with the same RBC judge.
Intermediate rewards measure progress against the rubric.
Because even useful searches and sufficient evidence can still yield an incomplete plan, an outcome judge separately evaluates the final video prompt and selected images against the request.
The process judge scores on the range $[0,2]$. To improve the accuracy of the outcome judge, we set its scoring range to $[0,1]$ and multiply its output credit by 2 to align the two complementary reward ranges.

For rollout $g$, let $r_{g,t}$ be the process credit, $o_g$ the outcome credit, and $T_g$ its number of assistant turns.
We assign
\begin{equation}
 R_{g,t}=\begin{cases}
 r_{g,t}, & t<T_g,\\
 \lambda r_{g,T_g}+(1-\lambda)o_g, & t=T_g.
 \end{cases}
 \label{eq:reward}
\end{equation}
The terminal mixture evaluates both the planning decision and the completeness of the resulting generation input.

We normalize rewards across rollouts of the same prompt at the same assistant-turn position.
This grouping does not assume identical action semantics at a given turn; rather, it compares context-dependent decisions for the same request after equal interaction rounds.
RBC evaluates each decision against the remaining task requirements and available evidence, mapping heterogeneous actions to a common task-progress scale. 
This allows continued search and termination to be compared as alternative decisions about whether additional evidence is needed. 
Turn position serves as a coarse control for interaction progress, rather than a semantic alignment of actions.
Let $\mathcal G_t$ denote the rollouts containing a turn at position $t$.
The advantage is
\begin{equation}
 A_{g,t}=\frac{R_{g,t}-\mu_t}{\sigma_t+\epsilon},
 \label{eq:advantage}
\end{equation}
where $\mu_t$ and $\sigma_t$ are the mean and standard deviation of
$\{R_{g,t}:g\in\mathcal G_t\}$, and $\epsilon$ is a stabilizer.
When $|\mathcal G_t|=1$, we set the advantage to zero because no relative comparison is available; such turns are empirically less than 1\%.

We use immediate RBC rewards deliberately, because dense process supervision provides a direct assessment of each decision under its observed context. 
Propagating downstream rewards can confound a useful search action with later planning errors, or reinforce redundant actions that happen to precede a successful completion. 
We therefore supervise intermediate decisions through local process credit and assess the completed generation input through the terminal reward. This design prioritizes targeted process supervision over explicit return-based attribution.
Experiments on normalization and discounted returns in \cref{tab:ablation} compare different design choices.
Finally, GRPO applies $A_{g,t}$ to the assistant tokens of turn $t$, with tool observations excluded from policy supervision~\cite{shao2024deepseekmath}.
Neither process nor outcome scoring requires video generation, avoiding rendering-induced variation in the training rewards.

%% file: sec/4_results.tex
\begin{figure}[t]
\centering
\includegraphics[width=0.95\linewidth]{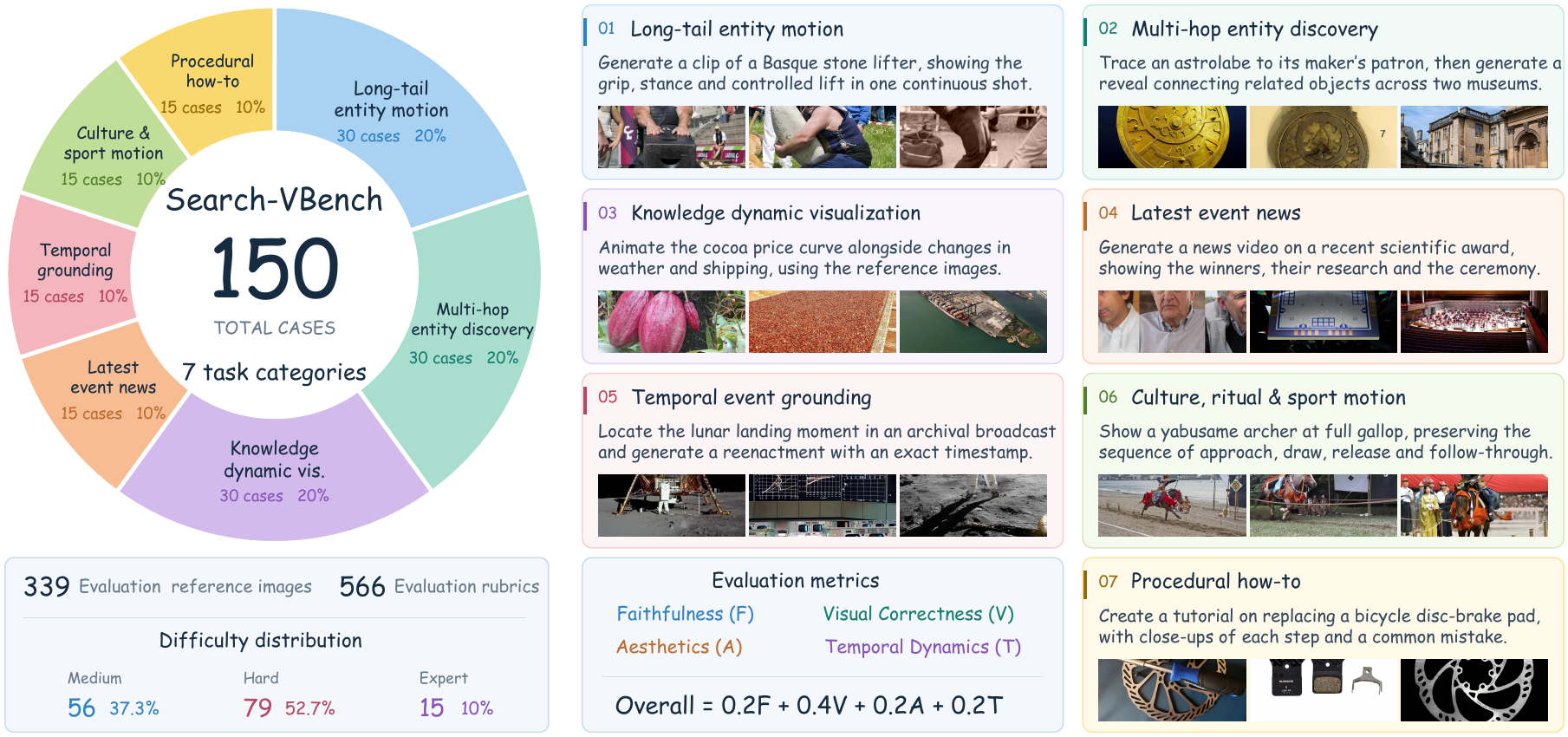}
\caption{\textbf{Search-VBench covers seven forms of knowledge-intensive video generation.}
The gallery summarizes representative requests, category counts, reference images, evaluation rubrics, difficulty levels, and the four dimensions used to score generated videos.}
\label{fig:benchmark}
\end{figure}

\section{Experiments}
\label{sec:experiment}
\subsection{Setup}
\label{sec:setup}

\noindent \textbf{Training Details.}
We use Qwen3-VL-8B-Instruct~\cite{bai2025qwen3} as the agent backbone. RBC-SFT is trained on approximately 16K filtered teacher trajectories, and RBC-GRPO is initialized from this checkpoint using a separate pool of 2K prompts. GLM-5.3-Flash~\cite{zeng2026glm} serves as the process judge. We use AdamW with learning rates of $10^{-5}$ (SFT) and $10^{-6}$ (RL), and batch size 16. Each RL prompt has six rollouts, up to ten interaction turns, five returned images per turn, a 36K context length, and a 4K response limit per turn. Overlong and repetitive rollouts are masked. Unless otherwise specified, the final reward uses $\lambda=0.3$. Additional details on data construction, system prompts, judge prompts, and reproducibility are provided in the supplementary material.

\noindent \textbf{Search-VBench.}
Search-VBench comprises 150 held-out prompts across seven categories (\cref{fig:benchmark}), with 339 reference images and 566 evaluation rubric items. It evaluates Faithfulness ($F$), Visual Correctness ($V$), Temporal Dynamics ($T$), and Aesthetics ($A$), measuring content fidelity, knowledge-dependent appearances, action and event ordering, and visual quality, respectively. Gemini 3.1 Pro~\cite{gemini31pro} scores each dimension on a three-level scale ($0$, $0.5$, or $1$), with three independent evaluations per video. We average the evaluations and compute Overall as $0.2F+0.4V+0.2T+0.2A$. When no valid reference image is available, $V$ is omitted, and the remaining weights are renormalized. All reported Search-VBench scores are scaled to $[0,100]$. 

\noindent \textbf{External Benchmarks and Baselines.}
Sci-VBench~\cite{zhang2026sci} evaluates Visual Quality (VT), Spatiotemporal Consistency (SC), Prompt Grounding (PG), and Scientific and Causal Correctness (SCC); we report its native average score. KIVI-Bench~\cite{wang2026knowledge} evaluates 60-second videos in terms of Factual Precision (FactP), Relevance (Rel), Completeness (Cmp), Clarity (Clr), and Helpfulness (HelpS). Some results are re-evaluated using the official repository. We compare direct generation and Loupe under identical generator and inference settings. MiniMax-H3, Wan 3.0, and Seedance 2.0 are evaluated on Search-VBench and Sci-VBench, while KIVI-Bench uses MiniMax-H3 due to generation costs. Additional standalone generators serve as reference baselines. We also report Gemini 3.1 Pro as a strong teacher reference under the same evaluation harness. VideoGen-Agent is the closest learned agentic video generation system, but its closed-source implementation prevents direct comparison. Finally, we implement two Qwen3-VL-8B-Instruct-based agentic baselines: Prompt Rewriting, which expands and rewrites prompts, and Workflow, which uses Loupe's search-enabled agent harness without training.

\subsection{Main Results on Search-VBench}
\label{sec:main_results}
The results in \cref{tab:search_vbench} show that Loupe improves every Search-VBench dimension for all three paired backbones.
Overall rises by 37.87 points for MiniMax-H3, 34.20 for Wan 3.0, and 20.67 for Seedance 2.0.
The largest gains are in Visual Correctness, which increases by 55.33, 52.66, and 41.33 points, respectively.
\input{table/search_vbench}

These gains match the role of retrieved references: a named entity becomes a concrete visual target rather than an appearance inferred from the short request alone.
With MiniMax-H3, Loupe surpasses Prompt Rewriting and Workflow across all four dimensions, improving the overall score by 35.27 and 21.27 points, respectively. This confirms that training yields gains beyond prompt refinement and search alone, and Loupe's results are also close to those of the strong closed-source Gemini 3.1 Pro model.
The agent also improves temporal content without replacing the renderer.
MiniMax-H3 and Wan 3.0 gain 21.66 and 21.00 points in Temporal Dynamics, while Seedance 2.0 gains 5.00 points from a stronger baseline.
This convergence indicates that missing knowledge and planning account for part of the gap between their direct-generation results.
Additional experimental results, including per-category results,
human evaluation, and judge--human agreement, are provided in
\cref{sec:supp_experiments}.

\subsection{Performance on Sci-VBench and KIVI-Bench}
\label{sec:transfer}
As shown in \cref{tab:sci_vbench}, Loupe improves performance on scientific content on Sci-VBench.
The Avg. scores rise from 2.74 to 3.37 for MiniMax-H3, from 2.83 to 3.44 for Wan 3.0, and from 3.22 to 3.67 for Seedance 2.0; in particular, Loupe enables Seedance 2.0 to surpass Gemini-Omni-Flash and become the new state of the art among our reproduced baselines.
VT changes by at most 0.04 across the paired backbones, so the gains primarily concern what the video depicts rather than low-level visual quality.
\input{table/sci_vbench}

The KIVI-Bench results in \cref{tab:kivi_bench} show that Loupe also improves the factuality and usefulness of MiniMax-H3's long-form videos.
FactP rises from 74.1 to 83.1 and HelpS from 50.9 to 63.5.
Details of the reproduction experiments are provided in \cref{sec:supp_evaluation}.

\input{table/kivi_bench}

\FloatBarrier
\subsection{Ablation Studies}
\label{sec:ablation}
We ablate Loupe on Search-VBench with MiniMax-H3
(\cref{tab:ablation} and~\cref{tab:supp_rubric_ablation}).
Base passes the original prompt directly to the generator.
All RL variants use the same training data and number of updates.

\textbf{(1) Demonstration weighting.}~Vanilla SFT raises Overall from 32.80 to 62.00.
RBC-SFT adds 2.47 points by emphasizing useful decisions within each demonstration, such as reducing redundant search and terminating early.

\textbf{(2) Step-level credit.}~RBC-GRPO ($\lambda=0.3$) consistently outperforms Traj-GRPO under both initializations: by 1.73 points with Vanilla SFT and 4.54 points with RBC-SFT.
This demonstrates the benefits of step-specific feedback beyond trajectory-level rewards.

\textbf{(3) Complete system comparison.}~Our full training pipeline, combining RBC-SFT and RBC-GRPO ($\lambda=0.3$),
outperforms conventional SFT followed by Traj-GRPO by 5.27 Overall points
(70.67 vs. 65.40).
This demonstrates the effectiveness of integrating demonstration-level weighting
and step-level credit assignment into a unified training framework.

\input{table/ablation}

\textbf{(4) Terminal reward.}~The best tested mixture ($\lambda=0.3$) reaches 70.67 Overall,
compared with 68.87 for process credit alone.
The final-plan reward checks whether useful evidence gathered across turns
is assembled into a complete plan.

\Needspace{2\baselineskip}
\textbf{(5) Reward propagation.}~At $\lambda=0.3$, immediate credit exceeds discounted return
($\gamma=0.9$) by 3.07 Overall points.
Immediate credit evaluates each search decision directly,
limiting the influence of later planning errors on its training signal.

\textbf{(6) Reward normalization.}~Normalizing at each turn position exceeds pooling by final versus
non-final turns by 2.74 Overall points.
Position-wise grouping compares rollouts after the same number of
interaction rounds, providing a reference for how much progress
is expected at that point.

\textbf{(7) Rubric specificity.}~Prompt-specific rubrics improve Overall by 3.74 points over a fixed rubric
and 5.40 points over scoring without a rubric.
They give the judge concrete criteria for whether a step addresses
the facts, appearances, or event order required by the request.

\input{table/ablation_rubric}

\subsection{RL Training Dynamics}
\label{sec:analysis}
\begin{figure}[htbp]
\centering
\includegraphics[width=0.85\linewidth]{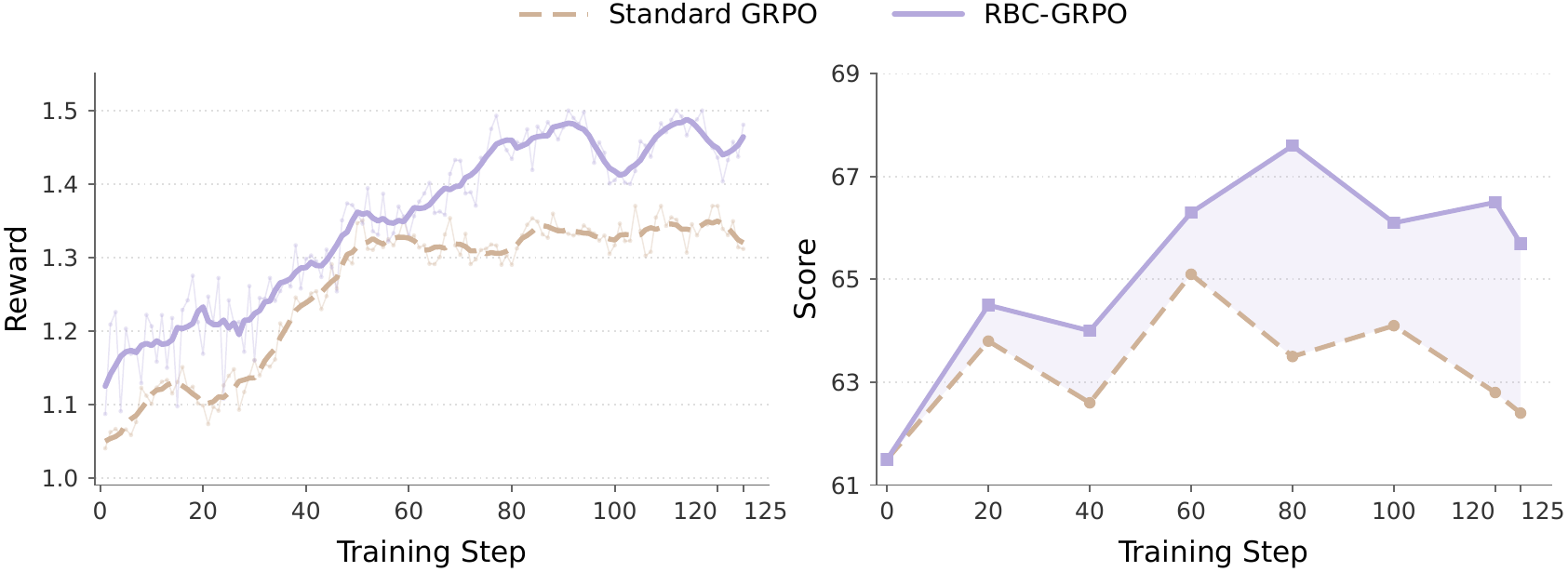}
\caption{\textbf{RL training dynamics.} The complete curves show training reward on the left and evaluation score on the right.}
\label{fig:grpo_reward_score}
\end{figure}

The curves in \cref{fig:grpo_reward_score} show that RBC-GRPO reaches higher training rewards and scores than trajectory-level GRPO.
The left panel shows the reward during optimization, and the right panel evaluates checkpoints on 50 difficult prompts constructed independently.
RBC-GRPO scores higher at every displayed noninitial evaluation, consistent with the full benchmark advantage over the baseline in \cref{tab:ablation}.

\subsection{Qualitative Comparisons}
\label{sec:qualitative}
\begin{figure}[H]
\centering
\includegraphics[width=\linewidth]{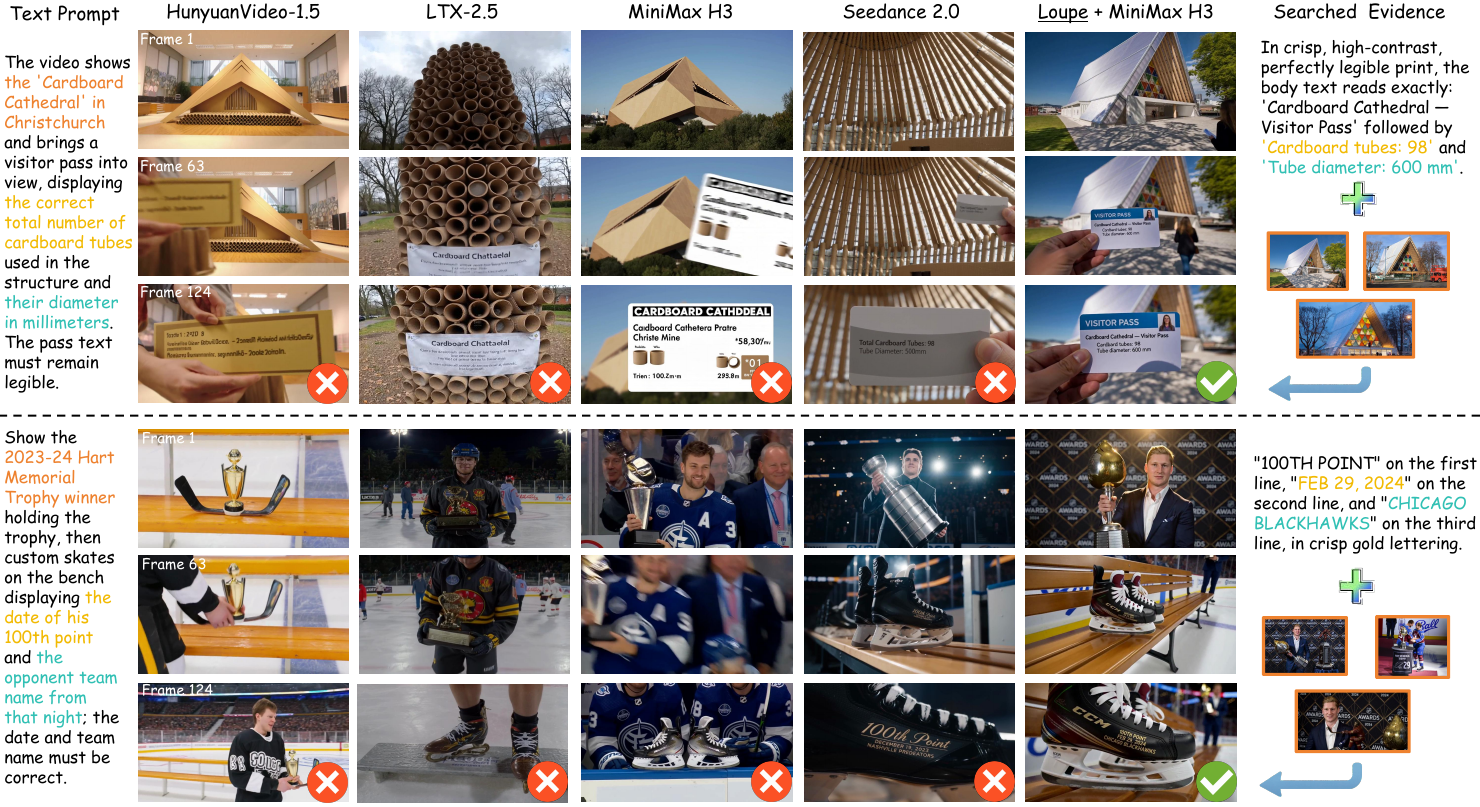}
\caption{\textbf{Qualitative comparisons.}
In the text prompt, \textcolor{reqbrown}{brown}, \textcolor{reqyellow}{yellow}, and \textcolor{reqcyan}{cyan} highlight distinct requirements involving visual and factual knowledge.
Existing generators hallucinate when knowledge gaps prevent them from satisfying these requirements.
As the search trajectory shows, \textbf{Loupe} retrieves relevant visual and factual evidence and thereby enables the generator to produce accurate results that satisfy the specified requirements.
Complete search trajectories for Loupe are provided in \cref{sec:supp_trajectories}.
}
\label{fig:qualitative_comparison}
\end{figure}

As shown in \cref{fig:qualitative_comparison}, existing generators hallucinate when knowledge gaps prevent them from satisfying these requirements.
By retrieving relevant visual and factual evidence, Loupe enables the video generator to produce accurate results that satisfy the specified grounding requirements.

%% file: table/search_vbench.tex
\begin{table}[t]
\centering
\caption{\textbf{Loupe improves all four Search-VBench dimensions across three backbones.} Green arrows denote absolute gains over each backbone. Gemini 3.1 Pro serves as a strong teacher reference. \textbf{Bold} marks the best value in each group, excluding Gemini.}
\label{tab:search_vbench}
\small
\setlength{\tabcolsep}{4pt}
\setlength{\aboverulesep}{0.3ex}
\setlength{\belowrulesep}{0.3ex}
\renewcommand{\arraystretch}{0.90}
\resizebox{\textwidth}{!}{%
\begin{tabular}{lccccc}
\toprule
\textbf{Method}
& \textbf{\makecell{Faithfulness $\uparrow$}}
& \textbf{\makecell{Visual\\Correctness $\uparrow$}}
& \textbf{\makecell{Temporal\\Dynamics $\uparrow$}}
& \textbf{\makecell{Aesthetics $\uparrow$}}
& \textbf{Overall $\uparrow$} \\
\midrule

\rowcolor{red4!20}
\multicolumn{6}{c}{\textit{\textbf{Open-source models}}} \\
\midrule
CogVideoX1.5-5B~\cite{yang2025cogvideox}
& 26.67 & 8.00 & 29.33 & 46.67 & 23.73 \\

HunyuanVideo-1.5~\cite{wu2025hunyuanvideo}
& 25.33 & \textbf{8.67} & 45.33 & 51.00 & 27.80 \\

LTX-2.5~\cite{hacohen2025ltx2}
& 23.33 & \textbf{8.67} & 45.33 & 53.00 & 27.80 \\

Wan 2.1 T2V-14B~\cite{wan21}
& 28.67 & 8.00 & 40.67 & 53.33 & 27.73 \\

Wan 2.2 T2V-A14B~\cite{wan22}
& 34.67 & \textbf{8.67} & 42.67 & 53.00 & 29.53 \\

MiniMax-H3~\cite{minimaxh3}
& \textbf{37.00} & 8.00 & \textbf{53.67}
& \textbf{57.33} & \textbf{32.80} \\

\midrule
\rowcolor{blue4!20}
\multicolumn{6}{c}{\textit{\textbf{Closed-source models}}} \\
\midrule
Wan 3.0~\cite{wan30}
& 48.00 & 9.67 & 55.67 & 59.00 & 36.40 \\

Seedance 2.0~\cite{seedance2026seedance}
& \textbf{65.67} & \textbf{20.00} & \textbf{69.33}
& \textbf{69.67} & \textbf{48.93} \\

\midrule
\rowcolor[HTML]{E6E6FA}
\multicolumn{6}{c}{\textit{\textbf{Agentic models}}} \\
\midrule
\textcolor{gray}{Gemini 3.1 Pro + MiniMax-H3}
& \textcolor{gray}{84.33}
& \textcolor{gray}{63.67}
& \textcolor{gray}{77.33}
& \textcolor{gray}{71.67}
& \textcolor{gray}{72.13} \\

Prompt Rewriting + MiniMax-H3
& 39.00 & 8.00 & 59.67 & 62.33 & 35.40 \\

Workflow + MiniMax-H3
& 60.33 & 29.33 & 61.33 & 66.67 & 49.40 \\

Loupe + MiniMax-H3 (Ours)
& \textbf{81.33}\gain{44.33}
& \textbf{63.33}\gain{55.33}
& 75.33\gain{21.66}
& 70.00\gain{12.67}
& \textbf{70.67}\gain{37.87} \\

Loupe + Wan 3.0 (Ours)
& 79.67\gain{31.67}
& 62.33\gain{52.66}
& \textbf{76.67}\gain{21.00}
& \textbf{72.00}\gain{13.00}
& 70.60\gain{34.20} \\

Loupe + Seedance 2.0 (Ours)
& 80.33\gain{14.66}
& 61.33\gain{41.33}
& 74.33\gain{5.00}
& 70.67\gain{1.00}
& 69.60\gain{20.67} \\

\bottomrule
\end{tabular}%
}
\end{table}

%% file: table/sci_vbench.tex
\begin{table}[t]
\centering
\caption{\textbf{Loupe improves scientific grounding on Sci-VBench.} \textbf{Bold} marks the best score within each group. Results for open-source and closed-source models are reproduced by us.}
\label{tab:sci_vbench}
\fontsize{9}{11}\selectfont
\setlength{\tabcolsep}{3pt}
\renewcommand{\arraystretch}{1}
\setlength{\aboverulesep}{0.3ex}
\setlength{\belowrulesep}{0.3ex}
\begin{tabular*}{\linewidth}{@{\extracolsep{\fill}}lccccc@{}}
\toprule
\textbf{Method}
& \textbf{VT $\uparrow$}
& \textbf{SC $\uparrow$}
& \textbf{PG $\uparrow$}
& \textbf{SCC $\uparrow$}
& \textbf{Avg. $\uparrow$} \\
\midrule

\rowcolor{red4!20}
\multicolumn{6}{c}{\textit{\textbf{Open-source models}}} \\
\midrule
CogVideoX1.5~\cite{yang2025cogvideox}
& 3.60 & 2.23 & 2.10 & 1.46 & 2.35 \\
LTX-2.3~\cite{hacohen2025ltx2}
& 3.66 & 2.35 & 2.15 & 1.44 & 2.40 \\
LongCat-Video~\cite{team2025longcat}
& 3.56 & 2.66 & 2.04 & 1.35 & 2.40 \\
LTX-2~\cite{hacohen2025ltx2}
& 3.76 & 2.76 & 1.80 & 1.31 & 2.41 \\
HunyuanVideo-1.5~\cite{wu2025hunyuanvideo}
& 3.48 & 2.90 & 2.08 & 1.47 & 2.48 \\
Wan2.2-5B~\cite{wan22}
& 3.51 & \textbf{2.98} & 2.15 & 1.39 & 2.51 \\
Cosmos3-Nano~\cite{agarwal2026cosmos}
& \textbf{3.77} & 2.92 & 2.03 & 1.40 & 2.53 \\
MiniMax-H3~\cite{minimaxh3}
& 3.64 & 2.94 & \textbf{2.64}
& \textbf{1.75} & \textbf{2.74} \\

\midrule
\rowcolor{blue4!20}
\multicolumn{6}{c}{\textit{\textbf{Closed-source models}}} \\
\midrule
Wan-2.6~\cite{wan26}
& 3.60 & 2.85 & 2.79 & 1.86 & 2.78 \\
Wan 3.0~\cite{wan30}
& 3.82 & 2.80 & 2.81 & 1.88 & 2.83 \\
Kling-2.6~\cite{kling26}
& 3.65 & \textbf{3.07} & 2.89 & 1.91 & 2.88 \\
Sora-2~\cite{sora2}
& 3.52 & 2.51 & 3.35 & 2.65 & 3.01 \\
Veo-3.1-Fast~\cite{veo3}
& 3.78 & 2.80 & 3.11 & 2.71 & 3.10 \\
Veo-3.1~\cite{veo3}
& 3.88 & 2.66 & 3.17 & 2.87 & 3.14 \\
Seedance 2.0~\cite{seedance2026seedance}
& 3.57 & 2.79 & 3.40 & 3.10 & 3.22 \\
HappyHorse-1.1~\cite{happyhorse}
& 3.71 & 3.07 & 3.55 & 2.94 & 3.32 \\
Gemini-Omni-Flash~\cite{geminiomni}
& \textbf{3.90} & 2.63 & \textbf{3.65}
& \textbf{3.38} & \textbf{3.39} \\

\midrule
\rowcolor[HTML]{E6E6FA}
\multicolumn{6}{c}{\textit{\textbf{Agentic models}}} \\
\midrule
Loupe + MiniMax-H3
& 3.67\gain{0.03}
& 3.33\gain{0.39}
& 3.51\gain{0.87}
& 2.97\gain{1.22}
& 3.37\gain{0.63} \\
Loupe + Wan 3.0
& \textbf{3.81}\scidrop{0.01}
& 2.83\gain{0.03}
& 3.69\gain{0.88}
& 3.43\gain{1.55}
& 3.44\gain{0.61} \\
Loupe + Seedance 2.0
& 3.61\gain{0.04}
& \textbf{3.58}\gain{0.79}
& \textbf{3.93}\gain{0.53}
& \textbf{3.56}\gain{0.46}
& \textbf{3.67}\gain{0.45} \\

\bottomrule
\end{tabular*}
\end{table}

%% file: table/kivi_bench.tex
\begin{table}[htbp]
\centering
\caption{\textbf{Loupe improves the factual precision and helpfulness of MiniMax-H3 on KIVI-Bench.} \textbf{Bold} marks the best value within each group. * denotes results we reproduced.}
\label{tab:kivi_bench}
\fontsize{9}{11}\selectfont
\setlength{\tabcolsep}{4pt}
\renewcommand{\arraystretch}{1}
\setlength{\aboverulesep}{0.3ex}
\setlength{\belowrulesep}{0.3ex}
\begin{tabular*}{\linewidth}{@{\extracolsep{\fill}}lccccccc@{}}
\toprule
& \multicolumn{3}{c}{\textbf{Factuality}}
& \multicolumn{4}{c}{\textbf{Helpfulness}} \\
\cmidrule(lr){2-4} \cmidrule(lr){5-8}
\textbf{Model}
& \textbf{\#Cor. $\uparrow$} & \textbf{Total} & \textbf{FactP $\uparrow$}
& \textbf{Rel $\uparrow$} & \textbf{Cmp $\uparrow$} & \textbf{Clr $\uparrow$} & \textbf{HelpS $\uparrow$} \\
\midrule

\rowcolor[HTML]{F5F5F5}
Human (reference)
& 426 & 437 & 97.8 & 84.0 & 78.7 & 83.2 & 81.9 \\

\midrule
\rowcolor{red4!20}
\multicolumn{8}{c}{\textit{\textbf{Open-source models}}} \\
\midrule
Wan 2.2~\cite{wan22}
& 306 & 426 & 73.1 & 57.4 & 53.0 & 34.8 & 48.4 \\
HunyuanVideo-1.5~\cite{wu2025hunyuanvideo}
& 259 & 403 & 63.2 & 41.3 & 38.0 & 19.3 & 32.9 \\
Helios-Base~\cite{yuan2026helios}
& 221 & 350 & 64.2 & 43.3 & 24.1 & 13.5 & 27.0 \\
LongCat-Video~\cite{team2025longcat}
& 184 & 341 & 50.8 & 23.3 & 15.4 & 7.2 & 15.3 \\
LongLive 1.0~\cite{yang2025longlive}
& 176 & 384 & 46.5 & 28.3 & 15.4 & 3.9 & 15.9 \\
MiniMax-H3*~\cite{minimaxh3}
& \textbf{388} & \textbf{528} & \textbf{74.1}
& \textbf{60.2} & \textbf{55.6} & \textbf{37.2} & \textbf{50.9} \\

\midrule
\rowcolor{blue4!20}
\multicolumn{8}{c}{\textit{\textbf{Closed-source models}}} \\
\midrule
HappyHorse 1.0~\cite{happyhorse}
& 362 & 437 & \textbf{83.2}
& 70.2 & 66.5 & 48.1 & 61.6 \\
Seedance 2.5~\cite{seedance25}
& \textbf{396} & \textbf{486} & 82.3
& \textbf{77.2} & \textbf{70.9} & \textbf{61.7} & \textbf{69.9} \\

\midrule
\rowcolor[HTML]{E6E6FA}
\multicolumn{8}{c}{\textit{\textbf{Agentic models}}} \\
\midrule
Loupe + MiniMax-H3 (Ours)
& \textbf{442} & \textbf{538} & \textbf{83.1}\gain{9.0}
& \textbf{70.6} & \textbf{66.7} & \textbf{53.3}
& \textbf{63.5}\gain{12.6} \\

\bottomrule
\end{tabular*}
\end{table}

%% file: table/ablation.tex
\begin{table}[htbp]
\centering
\caption{\textbf{Ablation Study.} 
\textbf{Bold} marks the best in each column.}
\label{tab:ablation}
\fontsize{9}{11}\selectfont
\setlength{\tabcolsep}{2.5pt}
\renewcommand{\arraystretch}{1.05}
\begin{tabular*}{\linewidth}{@{\extracolsep{\fill}}lccccc@{}}
\toprule
\textbf{Method} & \textbf{Faith.} & \textbf{Visual}
& \textbf{Temp.} & \textbf{Aesth.} & \textbf{Overall} \\
\midrule
Base
& 37.00 & 8.00 & 53.67 & 57.33 & 32.80 \\
\midrule
Vanilla SFT
& 76.00 & 45.67 & 71.67 & 71.00 & 62.00 \\
\quad + Traj-GRPO
& 79.33 & 52.33 & 73.33 & 69.67 & 65.40 \\
\quad + RBC-GRPO ($\lambda=0.3$)
& 80.33 & 55.67 & 73.67 & 70.33 & 67.13 \\
\midrule
RBC-SFT
& 79.00 & 49.33 & \textbf{76.00} & 68.67 & 64.47 \\
\quad + Traj-GRPO
& 80.00 & 53.67 & 74.33 & 69.00 & 66.13 \\
\quad + RBC-GRPO (discounted return, $\gamma=0.9$)
& 81.67 & 56.33 & 74.00 & 69.67 & 67.60 \\
\quad + RBC-GRPO (group normalization)
& 82.33 & 56.67 & 73.67 & 70.33 & 67.93 \\
\quad + RBC-GRPO ($\lambda=0.7$)
& \textbf{82.67} & 58.00 & 72.67 & \textbf{71.33} & 68.53 \\
\quad + RBC-GRPO ($\lambda=1.0$)
& 80.33 & 61.00 & 72.67 & 69.33 & 68.87 \\
\quad \textbf{+ RBC-GRPO ($\lambda=0.3$)}
& 81.33 & \textbf{63.33} & 75.33 & 70.00 & \textbf{70.67} \\
\bottomrule
\end{tabular*}
\end{table}

%% file: table/ablation_rubric.tex
\begin{table}[htbp]
\centering\small
\caption{\textbf{Rubric ablation across SFT and RL.}
\emph{No Rubric} scores intermediate steps without an explicit rubric;
\emph{Fixed Rubric} uses the same generic rubric for all prompts.}
\label{tab:supp_rubric_ablation}
\setlength{\tabcolsep}{5pt}
\begin{tabular}{lccccc}
\toprule
\textbf{Scoring context} & \textbf{Faith.} & \textbf{Visual}
& \textbf{Temp.} & \textbf{Aesth.} & \textbf{Overall} \\
\midrule
No Rubric
& 76.33 & 53.00 & 71.33 & \textbf{72.67} & 65.27 \\
Fixed Rubric
& \textbf{83.67} & 54.33 & 73.33 & 69.00 & 66.93 \\
\textbf{Prompt-Specific Rubric}
& 81.33 & \textbf{63.33} & \textbf{75.33} & 70.00
& \textbf{70.67} \\
\bottomrule
\end{tabular}
\end{table}

%% file: sec/5_conclusions.tex
\section{Conclusion}
\label{sec:conclusion}

We presented \textbf{Loupe}, an evidence-driven video generation agent that retrieves and integrates external textual and visual evidence to address knowledge gaps in video generation. To enable effective agent learning, we introduced \textbf{Rubric-Based Credit (RBC)}, which evaluates intermediate search and planning decisions against prompt-specific factual, visual, and temporal requirements. Building on RBC, we developed \textbf{RBC-SFT} and \textbf{RBC-GRPO} to incorporate fine-grained process supervision into supervised and reinforcement learning without video rendering during RL. We also constructed \textbf{Search-VBench} and demonstrated consistent improvements across multiple video generators and external benchmarks, with ablations validating the effectiveness of RBC training. Overall, our work highlights the importance of external evidence acquisition, grounded temporal planning, and request-conditioned process supervision for advancing knowledge-intensive video generation.

%% file: sec/X_suppl.tex
\pdfbookmark[1]{Appendix}{appendix-overview}
\section*{Appendix}
\setcounter{figure}{0}
\setcounter{table}{0}
\setcounter{equation}{0}
\renewcommand{\thefigure}{S\arabic{figure}}
\renewcommand{\thetable}{S\arabic{table}}
\renewcommand{\theequation}{S\arabic{equation}}
\renewcommand{\theHfigure}{appendix.\arabic{figure}}
\renewcommand{\theHtable}{appendix.\arabic{table}}
\renewcommand{\theHequation}{appendix.\arabic{equation}}

The appendix covers the following topics.

\begin{itemize}
    \item \textbf{\cref{sec:supp_data}} describes training data construction and teacher trajectory collection.

    \item \textbf{\cref{sec:supp_evaluation}} details Search-VBench, evaluation metrics, and external benchmark protocols.

    \item \textbf{\cref{sec:supp_implementation}} presents training configurations and implementation details.

    \item \textbf{\cref{sec:supp_experiments}} reports additional experiments, including category-wise results, judge reliability, overhead, human evaluation, and qualitative comparisons.

    \item \textbf{\cref{sec:supp_trajectories}} presents complete trajectory case studies of successful and failed retrieval and planning.

    \item \textbf{\cref{sec:limitations}} discusses limitations and future directions.

    \item \textbf{\cref{sec:supp_prompts}} provides the prompts and annotation instructions used in our framework and evaluations.
\end{itemize}

\section{Training Data Construction}
\label{sec:supp_data}

\subsection{Seed Sources and Task Coverage}
We draw knowledge-intensive seed requests from Gen-Searcher~\cite{feng2026gen} and GenEvolve~\cite{chen2026genevolve}, and use entity information from EntiWeave~\cite{entiweave} to broaden coverage.
Next, we use Gemini 3.1 Pro~\cite{gemini31pro}, Qwen3.8-Flash-Next~\cite{qiu2026design}, and GLM-5.3~\cite{zeng2026glm} to rewrite or generate requests that involve explicit action, state changes, or temporal development.
The resulting task pool spans the seven categories in \cref{tab:supp_taxonomy}.
The categories organize the primary knowledge need; a request can involve several grounding types.

\begin{table}[htbp]
\centering
\small
\renewcommand{\arraystretch}{1.12}
\setlength{\tabcolsep}{6pt}
\setlength{\aboverulesep}{2pt}
\setlength{\belowrulesep}{2pt}

\caption{\textbf{Training task taxonomy.} Descriptions distinguish the knowledge to acquire from the dynamic content to express.}
\label{tab:supp_taxonomy}

\begin{tabularx}{\textwidth}{
    >{\raggedright\arraybackslash}p{0.21\textwidth}
    >{\raggedright\arraybackslash}X
    >{\raggedright\arraybackslash}X
    cc
}
\toprule
\rowcolor{taxonomyHeader}
Category & Knowledge target & Dynamic requirement & SFT \% & RL \% \\
\midrule

Long-tail entity motion
& Distinctive identities, appearances, and attributes
& Preserve identity during a requested action or interaction
& 25.5 & 25.9 \\

\rowcolor{taxonomyStripe}
Multi-hop entity discovery
& An entity or fact identified through linked clues
& Depict the discovered subject and carry resolved facts into the scene
& 21.3 & 19.9 \\

Knowledge dynamic visualization
& Scientific, technical, or quantitative knowledge
& Show a mechanism, changing state, or readable information reveal
& 21.2 & 20.6 \\

\rowcolor{taxonomyStripe}
Latest event news
& Facts associated with a recent or time-sensitive event
& Depict the event and its specified factual state
& 8.0 & 8.7 \\

Temporal event grounding
& Participants, dates, and relations tied to an event
& Preserve event context and chronological development
& 15.1 & 14.3 \\

\rowcolor{taxonomyStripe}
Cultural ritual and sport motion
& Culturally or technically specific practices
& Present recognizable gestures, techniques, and interactions
& 4.3 & 5.4 \\

Procedural how-to
& Tools, constraints, and procedural knowledge
& Show ordered operations and their observable outcomes
& 4.6 & 5.1 \\

\bottomrule
\end{tabularx}
\end{table}

\subsection{Dynamic Rewriting and Rubric Construction}
A seed supplies the subject and its knowledge dependency; rewriting adds observable temporal content while preserving the facts that must be discovered.
For example, a static description of a Chope phulkari becomes a display followed by a flip that reveals its reversible embroidery, ending on a museum tag with requested factual details.
The rewritten request asks for the missing information without supplying its answer.
The rewriting template is provided in \cref{sec:prompt_seed}.

Each request is paired with reference requirements and grounding types.
A requirement record contains a target, a reason explaining its relevance to generation, and a priority.
These records separate what the request requires from any particular generated video.
Grounding labels specify the types of evidence needed to resolve those requirements.
The annotation vocabulary is described in \cref{tab:supp_grounding}.

\begin{table}[htbp]
\centering
\small
\caption{\textbf{Evaluation aspects for each grounding type.}}
\label{tab:supp_grounding}
\setlength{\tabcolsep}{5pt}
\renewcommand{\arraystretch}{1.12}
\setlength{\aboverulesep}{2pt}
\setlength{\belowrulesep}{2pt}

\begin{tabularx}{\columnwidth}{
    >{\raggedright\arraybackslash}p{0.35\columnwidth}
    >{\raggedright\arraybackslash}X
}
\toprule
\rowcolor{taxonomyHeader}
Grounding type & Meaning \\
\midrule

Text / numeric label
& Exact names, labels, quantities, units, or visible text \\

\rowcolor{taxonomyStripe}
Entity relationship
& Correct relations between requested entities \\

Entity appearance
& Recognizable identity and visual attributes \\

\rowcolor{taxonomyStripe}
Object / product details
& Model-specific components, features, or specifications \\

Historical accuracy
& Facts appropriate to the historical context \\

\rowcolor{taxonomyStripe}
Location / architecture
& Place identity, building form, and characteristic features \\

Multi-hop relation
& Linked evidence resolving an indirect description \\

\rowcolor{taxonomyStripe}
Temporal sequence
& Chronology or ordered event states \\

Latest factual state
& Facts dependent on the requested time or event \\

\rowcolor{taxonomyStripe}
Procedural order
& Ordering and dependencies of operations \\

Cultural practice
& Practice-specific objects, gestures, and conventions \\

\rowcolor{taxonomyStripe}
Scientific mechanism
& A scientific process and its relevant entities or states \\

Cause and effect
& The causal connection between intervention and outcome \\

\bottomrule
\end{tabularx}
\end{table}

\subsection{Teacher Rollouts, Filtering, and Splits}
The teacher~\cite{gemini31pro} interacts with the same text search, image search, and page-browsing environment as Loupe.
We retain the ordered assistant outputs, tool calls, returned observations, and final prompt with selected references.
This preserves the alignment between each assistant turn and the evidence available when it acts.
The Loupe system prompt in \cref{sec:prompt_system} specifies the interaction and output format.

Filtering removes trajectories with invalid formats, severe repetition, premature termination, or clear deviation from the task.
Local mistakes in an otherwise usable trajectory are handled by RBC rather than automatically triggering whole-trajectory rejection.

The retained SFT set contains approximately 16K trajectories.
The RL pool contains 2K prompts disjoint from the SFT prompts. The category distribution of the training data is given in \cref{tab:supp_taxonomy}, and Search-VBench is drawn from an independent held-out candidate pool.

\section{Search-VBench and Evaluation Details}
\label{sec:supp_evaluation}

\subsection{Composition and Reference Evidence}
The benchmark uses a held-out pool built through the knowledge-intensive request construction pipeline.
\cref{fig:benchmark} shows several qualitative examples.
The benchmark contains 339 reference-image entries and 566 evaluation rubrics. 
Difficulty annotations comprise 56 medium, 79 hard, and 15 expert requests.

This evaluation metadata is constructed with GLM-5.3-Flash~\cite{zeng2026glm} and then manually verified, with images that do not match the request filtered out.
These evaluation references serve a different role from images selected by the agent as generation inputs: the former define visual evidence for judging, while the latter condition rendering.

\subsection{Metric Interpretation and Aggregation}
The main paper gives the score scale and weights.
Here we clarify the boundary between dimensions: Faithfulness concerns compliance with requested subjects and content; Visual Correctness concerns knowledge-dependent identity and appearance relative to reference evidence.
Temporal Dynamics evaluates requested actions, ordering, continuity, and physical plausibility; Aesthetics evaluates presentation quality and visual artifacts.
An attractive but incorrectly identified entity can therefore receive a strong aesthetic assessment while failing visual grounding.
The evaluator template in \cref{sec:prompt_evaluation} makes these distinctions explicit.

For video $i$, dimension $d$, and independent judge call $k$, let $s_{i,d}^{(k)}\in\{0,0.5,1\}$.
We first average the three calls for that video and dimension.
Writing $m_{i,d}=0$ only for Visual Correctness when video $i$ lacks a valid evaluation reference, and $m_{i,d}=1$ otherwise, the dataset score is
\begin{equation}
 \mathrm{Overall}=\frac{100}{N}\sum_{i=1}^{N}
 \frac{\sum_d w_d m_{i,d}\bar s_{i,d}}
 {\sum_d w_d m_{i,d}}.
 \label{eq:supp_aggregate}
\end{equation}
Normalization is applied before averaging videos, so weighting displayed dimension means need not recover Overall when validity masks differ.

\subsection{External Benchmarks}
Following the original papers, we evaluate on the TestMini set for Sci-VBench~\cite{zhang2026sci} and on the same sampled subset of 54 prompts for KIVI-Bench~\cite{wang2026knowledge}.
For the reproduced results of Sci-VBench and KIVI-Bench in \cref{tab:sci_vbench} and~\cref{tab:kivi_bench}, we download the videos publicly released by the original authors and re-evaluate them with their open-source code. 
These results may differ somewhat from those reported in the original papers, but we use a consistent evaluation protocol for all our experiments on Loupe.
Because KIVI-Bench requires generating 60-second videos, we first use Loupe to obtain the evidence-grounded video prompt and reference images, and then use the official repository to synthesize the full long video step by step.

\subsection{Definitions of Ablation Baselines}
\label{sec:ablation_baselines}

We clarify the RL baselines and alternative reward designs in \cref{tab:ablation}. 
All variants use the same rollout prompts and training configurations, differing only in reward construction or normalization.

\noindent \textbf{Trajectory-level GRPO (Traj-GRPO).}
Instead of assigning step-specific process credits, Traj-GRPO uses only the final outcome reward $o_g$ for each rollout and broadcasts its trajectory-level advantage to all assistant turns. The advantage is normalized across rollouts of the same prompt.

\noindent \textbf{Discounted return.}
Instead of using immediate rewards $R_{g,t}$, this variant computes the discounted return:
\begin{equation}
G_{g,t}=\sum_{k=t}^{T_g}\gamma^{k-t}R_{g,k},
\end{equation}
where $\gamma=0.9$. The resulting returns are normalized across rollouts at corresponding turn positions.

\noindent \textbf{Terminal/non-terminal normalization.}
Instead of normalizing rewards separately at each turn position, this variant pools all intermediate turns into one group and final turns into another, computing group-relative advantages within each group. Our default position-wise normalization compares rollouts at the same interaction position, providing finer-grained progress references.

\noindent \textbf{Immediate credit.}
Our default approach uses $R_{g,t}$ directly for advantage estimation, without propagating downstream rewards. This isolates the supervision of each decision from subsequent planning outcomes.

\section{Training and Implementation Details}
\label{sec:supp_implementation}
\begin{table}[htbp]
\centering
\small
\caption{\textbf{Training settings for the SFT and RL stages.}}
\label{tab:supp_training}
\setlength{\tabcolsep}{5pt}
\renewcommand{\arraystretch}{1.12}
\setlength{\aboverulesep}{2pt}
\setlength{\belowrulesep}{2pt}

\begin{tabularx}{\columnwidth}{
    >{\raggedright\arraybackslash}p{0.43\columnwidth}
    >{\raggedright\arraybackslash}X
}
\toprule
\rowcolor{taxonomyHeader}
Setting & Value \\
\midrule

Agent backbone & Qwen3-VL-8B-Instruct \\

\rowcolor{taxonomyStripe}
SFT trajectories & 16K \\

RL prompt pool & 2K, disjoint from SFT \\

\rowcolor{taxonomyStripe}
Initialization for RL & RBC-SFT checkpoint \\

Process-credit judge & GLM-5.3-Flash \\

\rowcolor{taxonomyStripe}
Outcome judge & GLM-5.3-Flash \\

Optimizer, both stages & AdamW \\

\rowcolor{taxonomyStripe}
SFT / RL learning rate & $10^{-5}$ / $10^{-6}$ \\

Batch size, both stages & 16 \\

\rowcolor{taxonomyStripe}
Rollouts / group size & 6 per prompt \\

Final-turn mixture & $\lambda=0.3$ \\

\rowcolor{taxonomyStripe}
Maximum interaction turns & 10 \\

Returned images per turn & At most 5 \\

\rowcolor{taxonomyStripe}
Context / response budget & 36K / 4K per turn \\

Temperature / top-$p$ & 0.7 / 0.95 \\

\rowcolor{taxonomyStripe}
GRPO clip & 0.28 \\

SFT / RL epochs & 2 / 1 \\

\rowcolor{taxonomyStripe}
Hardware / precision & 16 A100 / BF16 \\

Seeds & 42 \\

\bottomrule
\end{tabularx}
\end{table}
\cref{tab:supp_training} summarizes the implementation settings.
The process and final-plan judge templates are provided in \cref{sec:prompt_scoring} and~\cref{sec:prompt_outcome}.
We use \href{https://serper.dev/}{Serper} for text and image search and \href{https://urltotext.com/}{URL to Text} for webpage browsing.
The tool-call limit per trajectory is 20 during teacher data collection, reduced to 10 during RL to mitigate training stalls, and restored to 20 for inference.
For video generation, all Search-VBench evaluations use 5-second videos at 480p resolution, while Sci-VBench and KIVI-Bench follow the configurations of their respective original papers.

\Needspace{6\baselineskip}
\section{Additional Experimental Results}
\label{sec:supp_experiments}
\subsection{Performance by Task Category}
\begin{table}[htbp]
\centering\small
\caption{\textbf{Evaluation results for Loupe + MiniMax-H3 across Search-VBench categories.}}
\label{tab:supp_category_results}
\setlength{\tabcolsep}{4pt}
\begin{tabularx}{\linewidth}{@{}>{\raggedright\arraybackslash}Xccccc@{}}
\toprule
\rowcolor[HTML]{E8F5EE}
Category & Faithfulness & \makecell{Visual\\Correctness} & \makecell{Temporal\\Dynamics} & Aesthetics & Overall\\
\midrule
Long-tail entity motion
& 85.00 & 78.33 & 80.00 & 73.33 & 79.00 \\

Multi-hop entity discovery
& 90.00 & 73.33 & 73.33 & 66.67 & 75.33 \\

Knowledge dynamic visualization
& 85.00 & 61.67 & 71.67 & 68.33 & 69.67 \\

Latest event news
& 73.33 & 43.33 & 80.00 & 66.67 & 61.33 \\

Cultural ritual and sport motion
& 90.00 & 50.00 & 80.00 & 83.33 & 70.67 \\

Temporal event grounding
& 70.00 & 60.00 & 73.33 & 70.00 & 66.67 \\

Procedural how-to
& 60.00 & 53.33 & 70.00 & 63.33 & 60.00 \\
\bottomrule
\end{tabularx}
\end{table}
\cref{tab:supp_category_results} decomposes Loupe + MiniMax-H3's Search-VBench performance.
Long-tail entity motion has the highest Overall score at 79.00, followed by multi-hop entity discovery at 75.33.
Procedural how-to and latest event news score 60.00 and 61.33, respectively.
The latter has the lowest Visual Correctness score, 43.33, despite a Temporal Dynamics score of 80.00.
This separates the ability to produce the requested dynamics from the ability to ground event-specific visual details.

\subsection{Quantifying RBC Judge Noise}
As RBC relies on an LLM-based process reward model, its scoring reliability may be affected by inherent judgment variability. 
Moreover, the adjustment ranges in \cref{tab:rubric_scoring} are manually specified, making it important to assess the stability and accuracy of the resulting credits.

We conduct two complementary evaluations. First, we randomly sample 1K trajectories and ask the RBC Judge to independently score each step three times. We define self-consistency as the proportion of steps whose maximum and minimum assigned credits differ by at most 0.1. Under this criterion, the judge achieves 93.6\% self-consistency, indicating stable scoring across repeated evaluations. Second, we randomly sample 100 steps for manual evaluation by five independent annotators. We measure Judge-human agreement as the proportion of comparisons in which the judge-assigned and human-assigned credits differ by at most 0.1. The resulting agreement reaches 91.8\% across 500 comparisons, further supporting the reliability of our rubric-based scoring scheme.

Overall, these results suggest that the RBC Judge provides stable and reasonably accurate step-level supervision, with limited sensitivity to repeated judgments.

\subsection{Training Overhead of RBC Scoring}
\textbf{Time Analysis}. The complete training pipeline consists of four stages: video prompt and rubric generation, teacher trajectory collection, RBC-SFT, and RBC-GRPO. Generating 25K video prompts and rubrics takes approximately 17 hours (20 s per prompt, concurrency 8), while collecting teacher trajectories takes approximately 70 hours (160 s per trajectory, concurrency 16). For RBC-SFT, scoring the collected trajectories requires approximately 13 hours (30 s per trajectory, concurrency 16), followed by 10 hours of training for two epochs. Thus, RBC scoring introduces an additional 13 hours compared with standard SFT. For RBC-GRPO, we asynchronously overlap step-level judging with subsequent response generation, keeping the parameter update time nearly identical to standard GRPO (700 s per update). Training for 125 steps takes approximately 25 hours.
Overall, the complete pipeline takes approximately 135 hours (5.6 days), with the additional RBC scoring overhead accounting for only 13 hours (9.6\%).

\textbf{Cost Analysis}. We further analyze the computational cost of RBC-based supervision. The RBC judge uses GLM-5.3-Flash for step-level scoring, Gemini 3.1 Pro is mainly used for prompt construction and teacher trajectory collection, and training incurs additional GPU cost.
The RBC judge accounts for less than 10\% of the overall pipeline cost, demonstrating that fine-grained process supervision introduces limited overhead.

\subsection{Human Evaluation and Judge Validation}
\label{sec:supp_human}

\noindent \textbf{Human preference.}
We conducted a side-by-side human evaluation on 300 randomly sampled, prompt-matched video pairs across all seven categories. For each backbone (MiniMax-H3, Wan 3.0, and Seedance 2.0), we evaluated 100 pairs comparing direct generation with Loupe-augmented generation.
Five raters independently assessed each pair, with model identities hidden and the left-right order randomized.
For each pair, raters selected their preferred video separately for Faithfulness, Visual Correctness, Temporal Dynamics, Aesthetics, and Overall quality, without a tie option.
This protocol yielded 500 rater-pair assessments per backbone and 1,500 assessments in total, each comprising five dimension-specific choices.

For each dimension, the Loupe preference rate was computed as the fraction of individual rater choices favoring Loupe (500 choices per backbone; 1,500 pooled).
As shown in \cref{tab:supp_human_preference}, Loupe was preferred across all five dimensions for every backbone.
The pooled Overall preference rate was 82.33\%, with backbone-specific rates of 89.00\% for MiniMax-H3, 85.00\% for Wan 3.0, and 73.00\% for Seedance 2.0.
Preference rates were highest for Visual Correctness (89.00\%) and Faithfulness (83.33\%), followed by Temporal Dynamics (75.33\%) and Aesthetics (68.33\%).

\begin{table}[htbp]
\centering\small
\setlength{\tabcolsep}{2.5pt}
\caption{\textbf{Human preference for Loupe-augmented generation (\%).}
$F$, $V$, $T$, and $A$ denote Faithfulness, Visual Correctness, Temporal Dynamics, and Aesthetics, respectively.
Overall is a separate holistic preference judgment.}
\label{tab:supp_human_preference}
\begin{tabular}{lccccc}
\toprule
\rowcolor[HTML]{E8F5EE}
Paired generator & $F$ & $V$ & $T$ & $A$ & Overall\\
\midrule
MiniMax-H3
& 92.00 & 94.00 & 80.00 & 71.00 & 89.00\\
\rowcolor[HTML]{F6F7F8}
Wan 3.0
& 86.00 & 91.00 & 79.00 & 72.00 & 85.00\\
Seedance 2.0
& 72.00 & 82.00 & 67.00 & 62.00 & 73.00\\
\midrule
\rowcolor[HTML]{F6F7F8}
All comparisons
& 83.33 & 89.00 & 75.33 & 68.33 & 82.33\\
\bottomrule
\end{tabular}
\end{table}

\noindent \textbf{Judge--human agreement.}
We used the same 300 video pairs to evaluate the agreement between Gemini 3.1 Pro and human preferences.
For each dimension, the majority vote of the five raters served as the human reference.
The automatic preference was determined by comparing the two videos' scores for the corresponding dimension; for Overall quality, we compared their automatically aggregated scores.
When the scores were tied, we made an additional pairwise Gemini call with both videos and the corresponding rubric, requiring a preference without a tie option.
For dimension $d$, we computed agreement as
\begin{equation}
 \mathrm{Agreement}_d=\frac{100}{300}
 \sum_{j=1}^{300}\mathbf{1}
 [\hat y_{j,d}=y^{\mathrm{majority}}_{j,d}],
\end{equation}
where $\hat y_{j,d}$ is the automatic preference and
$y^{\mathrm{majority}}_{j,d}$ is the human majority preference for pair $j$.

\begin{table}[htbp]
\centering\small
\setlength{\tabcolsep}{3pt}
\renewcommand{\arraystretch}{1}
\caption{\textbf{Gemini--human preference agreement (\%).}
$F$, $V$, $T$, and $A$ denote the same dimensions as in \cref{tab:supp_human_preference}.}
\label{tab:supp_judge_agreement}
\begin{tabular}{lccccc}
\toprule
\rowcolor[HTML]{E8F5EE}
Scope & $F$ & $V$ & $T$ & $A$ & Overall\\
\midrule
Long-tail entity motion
& 88.89 & 91.11 & 86.67 & 82.22 & 86.67\\
\rowcolor[HTML]{F6F7F8}
Multi-hop entity discovery
& 88.89 & 88.89 & 84.44 & 80.00 & 88.89\\
Knowledge dynamic visualization
& 85.71 & 90.48 & 85.71 & 78.57 & 88.10\\
\rowcolor[HTML]{F6F7F8}
Latest event news
& 88.10 & 90.48 & 85.71 & 80.95 & 88.10\\
Temporal event grounding
& 85.71 & 88.10 & 83.33 & 78.57 & 85.71\\
\rowcolor[HTML]{F6F7F8}
Cultural ritual and sport motion
& 88.10 & 90.48 & 85.71 & 80.95 & 88.10\\
Procedural how-to
& 88.10 & 90.48 & 88.10 & 80.95 & 88.10\\
\midrule
\rowcolor[HTML]{F6F7F8}
All categories
& 87.67 & 90.00 & 85.67 & 80.33 & 87.67\\
\bottomrule
\end{tabular}
\end{table}

\cref{tab:supp_judge_agreement} reports category-specific agreement.
Across all categories, Gemini achieved 87.67\% agreement with human majority preferences on Overall quality.
Dimension-specific agreement ranged from 80.33\% for Aesthetics to 90.00\% for Visual Correctness.
Overall agreement was also consistent across categories, ranging from 85.71\% to 88.89\%.
These results support the use of Gemini 3.1 Pro as an automatic judge under this evaluation protocol.

\subsection{Additional Qualitative Comparisons}
\begin{figure}[htbp]
\centering
\includegraphics[width=0.95\linewidth]{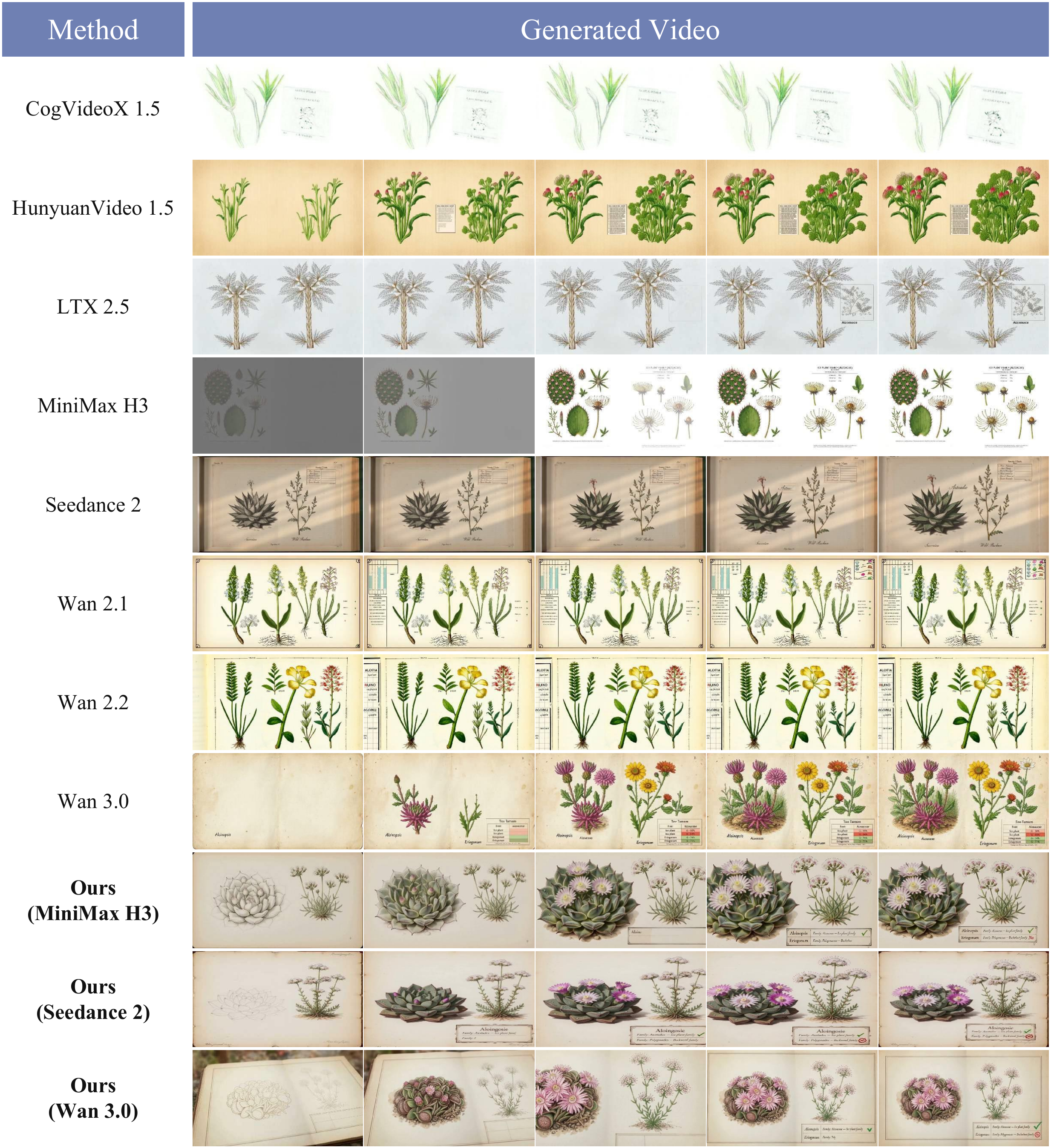}
\caption{\textbf{Qualitative comparison on Knowledge Dynamic Visualization.}
\textbf{Input prompt:} \textit{The video shows a botanical illustration plate coming to life: specimens of both Aloinopsis and Eriogonum are illustrated side by side with botanically correct morphological features, while a corner taxonomical chart updates to clearly indicate whether each belongs to the `ice plant' family (Aizoaceae).}
Each row shows five sampled frames from a video generated by a baseline model or our method. The task requires both botanical detail and an evolving taxonomic explanation. Baseline results often depict generic plant illustrations or weakly distinguish the requested specimens. Our results show a more consistent side-by-side composition of distinct specimens and more explicit taxonomic annotations as the illustrations develop.}
\label{fig:knowledge_dynamic_visualization_1985}
\end{figure}

\begin{figure}[htbp]
\centering
\includegraphics[width=0.95\linewidth]{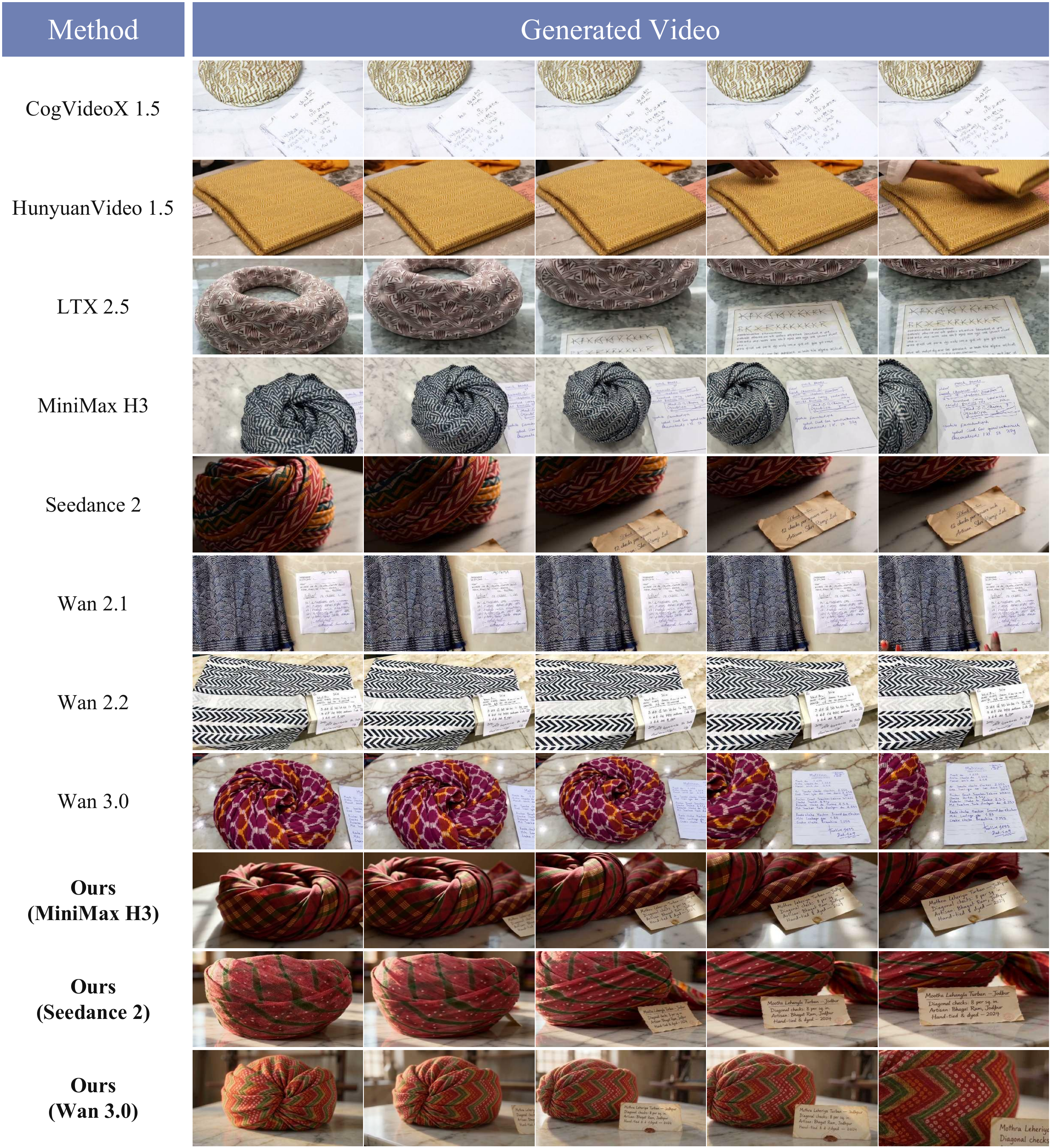}
\caption{\textbf{Qualitative comparison on Culture, Ritual, and Sport Motion.}
\textbf{Input prompt:} \textit{The video shows a close-up of a `Mothra' style leheria turban from Jodhpur on a polished marble surface, revealing its rhythmic intersecting wave patterns and diagonal `checks'. It also reveals a merchant's handwritten receipt beside the turban, where the specific number of diagonal `checks' per square inch and the local name of the artisan who pioneered this variant must remain legible and correct.}
Each row shows five uniformly sampled frames from a video generated by a baseline model or our method. The task couples a culturally specific textile pattern with a camera reveal of a handwritten receipt. Several baselines produce folded fabric, generic motifs, or poorly readable writing. Our results retain a wrapped turban and more clearly reveal the receipt, with explicit check-count and artisan-name fields.}
\label{fig:culture_ritual_sport_motion_1359}
\end{figure}

\begin{figure}[htbp]
\centering
\includegraphics[width=0.95\linewidth]{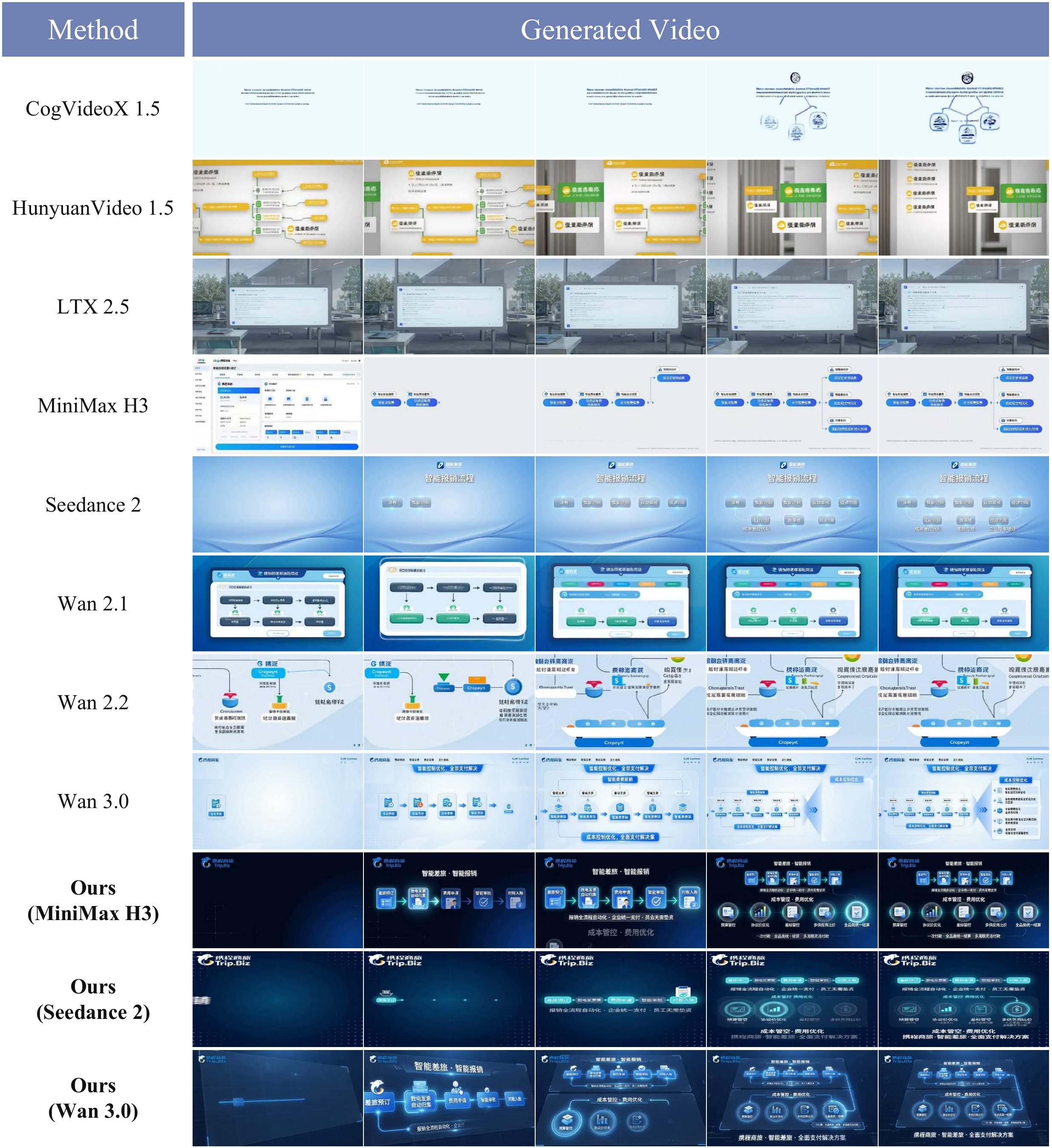}
\caption{\textbf{Qualitative comparison on Procedural How-To.}
\textbf{Input prompt:} \textit{Create a video of a modern corporate financial reimbursement interface for Ctrip Corporate Travel, in which an on-screen flowchart animates step by step through the key nodes of its intelligent expense reimbursement process, then highlights its cost control optimization goals. The process labels and the brand's visual identity, reflecting its intelligent reimbursement and comprehensive payment solutions, must remain correct and clearly legible throughout.}
Each row shows five uniformly sampled frames from a video generated by a baseline model or our method. The task requires an ordered workflow animation followed by a cost-control summary, with consistent branding and readable labels. Several baselines show generic dashboards, unstable layouts, or distorted text. Our results more clearly separate the reimbursement workflow from the subsequent optimization summary, using connected process nodes, progressive highlighting, and a persistent brand header.}
\label{fig:procedural_how_to_18017}
\end{figure}

\begin{figure}[htbp]
\centering
\includegraphics[width=0.95\linewidth]{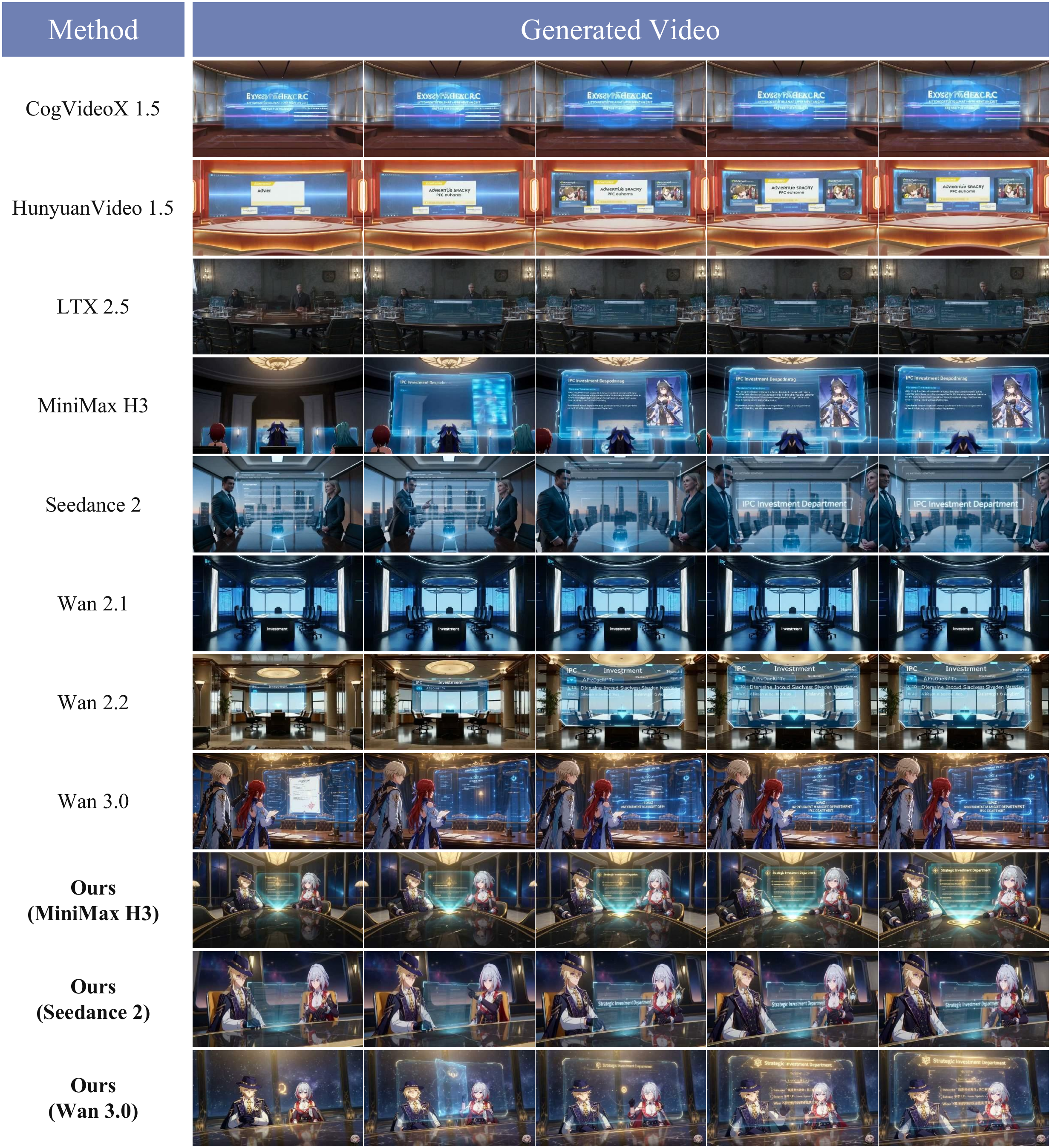}
\caption{\textbf{Qualitative comparison on Long-Tail Entity Motion.}
\textbf{Input prompt:} \textit{A luxury IPC boardroom becomes the setting as Aventurine and Topaz from Honkai: Star Rail review a digital contract. The holographic display must update to correctly show the official name of the IPC `Investment' department they both lead, with the label remaining legible.}
Each row shows five uniformly sampled frames from a video generated by a baseline model or our method. The task combines two named characters with a legible text update during their interaction with a digital contract. Several baselines reproduce the boardroom and holographic display but omit the characters, replace their appearances, or render distorted department labels. Our results preserve more distinctive hairstyle and costume cues for both characters while revealing a readable ``Strategic Investment Department'' heading on the display.}
\label{fig:long_tail_entity_motion_2168}
\end{figure}

Additional qualitative video comparisons are provided in \cref{fig:knowledge_dynamic_visualization_1985}--\cref{fig:long_tail_entity_motion_2168}, covering knowledge dynamic visualization, culture, ritual, and sport motion, long-tail entity motion, and procedural how-to generation. Each figure presents five sampled frames per video to illustrate both visual content and temporal progression. These examples examine complementary challenges: animating botanical illustrations with taxonomic annotations, revealing specific textile patterns and handwritten information, coordinating interactions among named characters, and presenting an ordered reimbursement workflow with readable interface labels.

Across these cases, baseline generators often reproduce the broad setting or action while missing finer requirements, such as distinctive character appearances, the requested number of participants, or legible task-specific text. Our results generally show clearer differentiation of the requested entities and a closer correspondence between the evolving scene and the information specified in the prompt. The comparisons also reveal remaining limitations: fine-grained attributes can still deviate from the request, as illustrated by the hair-color mismatch in \cref{fig:long_tail_entity_motion_2168}, and readable generated annotations alone do not establish factual correctness. Together, these examples illustrate both the improvements and the remaining challenges in maintaining entity-specific details and semantic constraints throughout video generation.

\section{Complete Trajectory Case Studies}
\label{sec:supp_trajectories}
\begin{figure}[htbp]
\centering
\includegraphics[width=\linewidth]{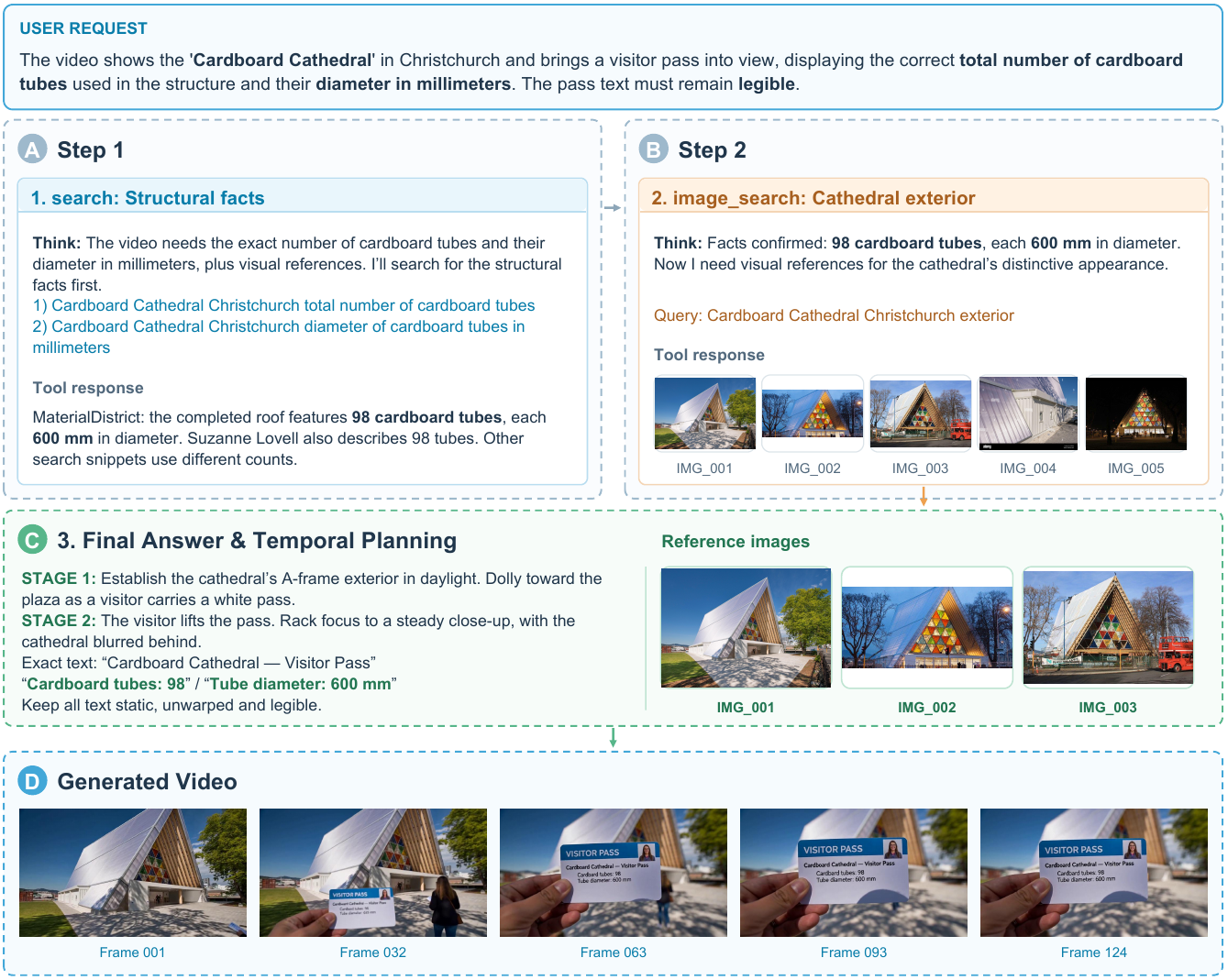}
\caption{\textbf{Case 1 of the complete inference trajectory of Loupe shown in \cref{fig:qualitative_comparison}.}}
\label{fig:supp_figure6_detail_1}
\end{figure}

\begin{figure}[htbp]
\centering
\includegraphics[width=0.7\linewidth]{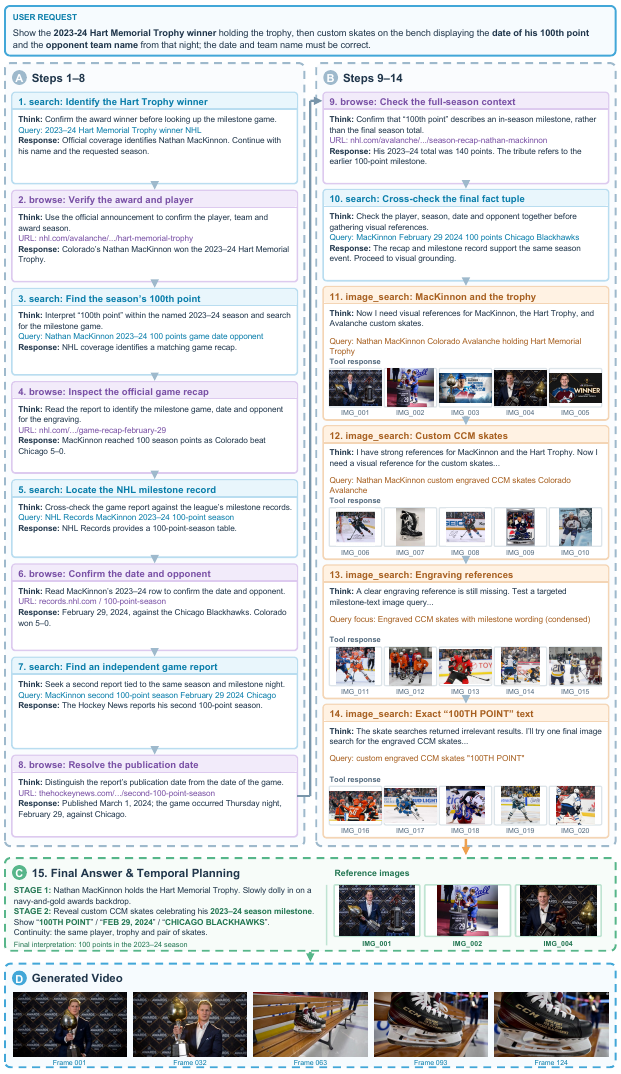}
\caption{\textbf{Case 2 of the complete inference trajectory of Loupe shown in \cref{fig:qualitative_comparison}.}}
\label{fig:supp_figure6_detail_2}
\end{figure}

\cref{fig:supp_figure6_detail_1} and~\cref{fig:supp_figure6_detail_2} show the complete reasoning process of Loupe from \cref{fig:qualitative_comparison}.

\begin{figure}[htbp]
\centering
\includegraphics[width=0.85\linewidth]{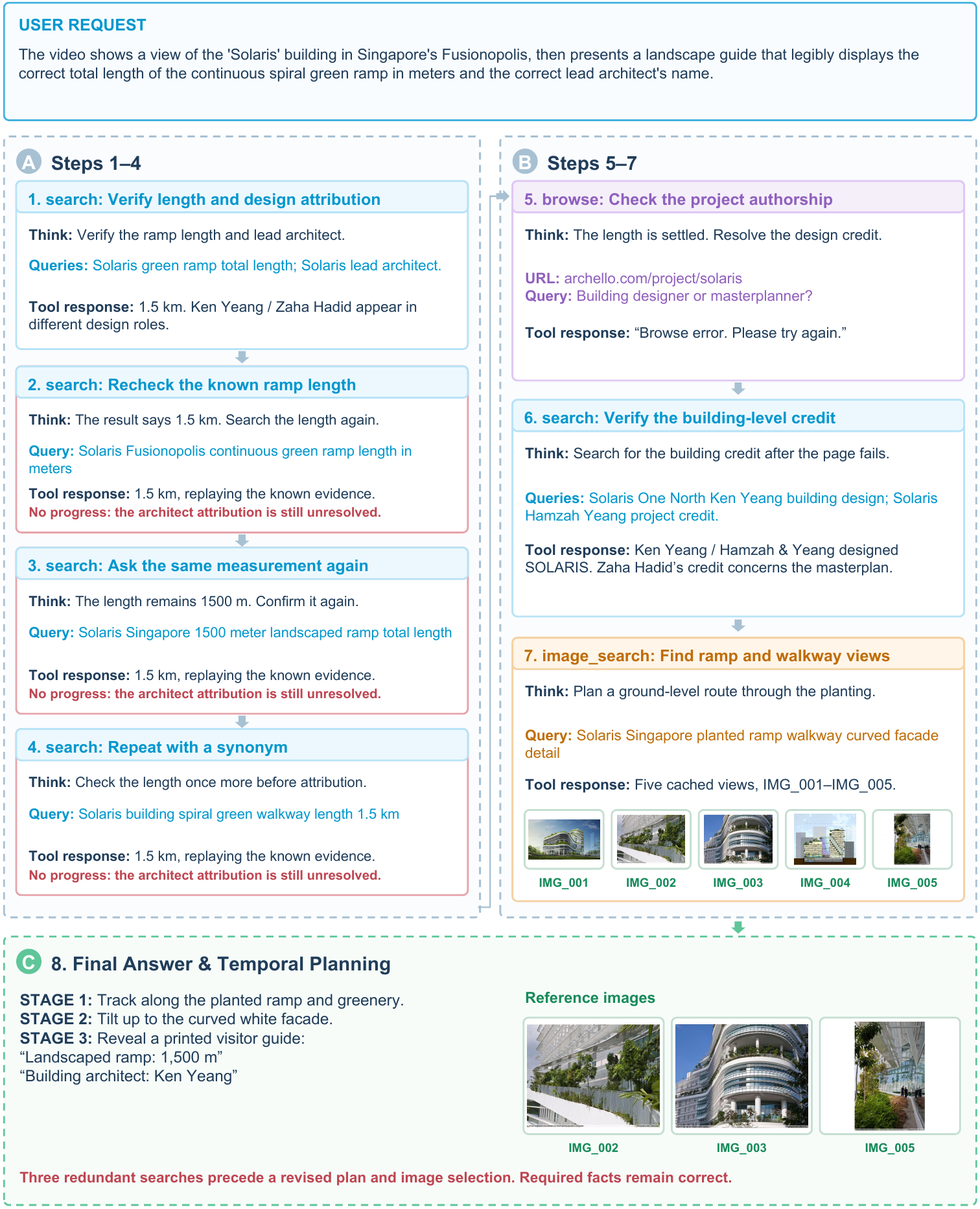}
\caption{
\textbf{Failure trajectory characterized by redundant retrieval.}
The agent repeatedly searches for the already established ramp length
while leaving the architect attribution unresolved.
Although subsequent steps produce a factually correct plan with different
reference images and camera movements, the repeated searches add no
task-relevant evidence and delay progress.
}
\label{fig:solaris_failure}
\end{figure}

\begin{figure}[htbp]
\centering
\includegraphics[width=0.85\linewidth]{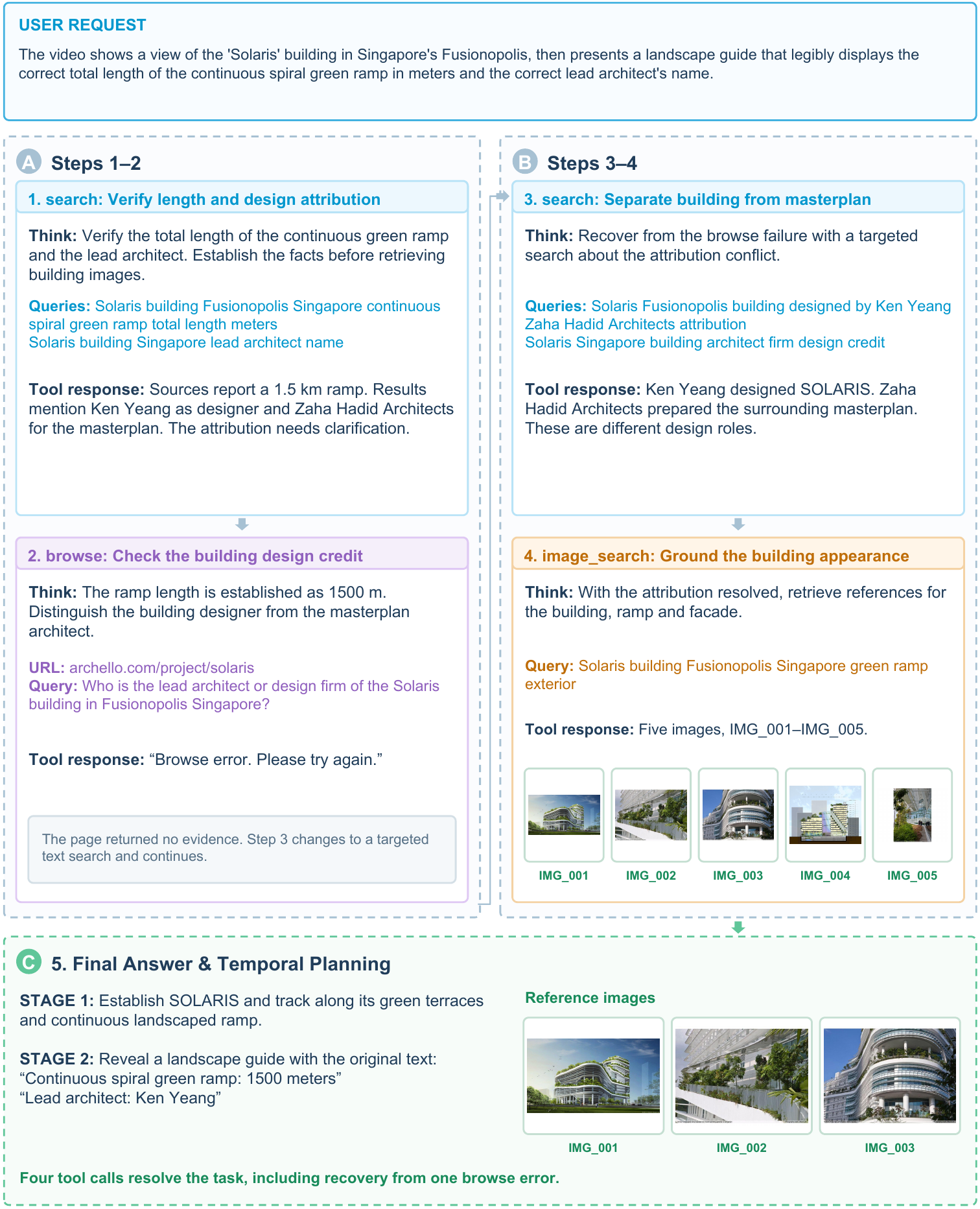}
\caption{
\textbf{Successful trajectory for grounded architectural planning.}
The agent resolves the distinction between the building designer and
the masterplan architect, recovers from a failed browse call through
targeted search, and selects relevant visual references to produce
a factually grounded video plan.
}
\label{fig:solaris_success}
\end{figure}

\begin{figure}[htbp]
\centering
\includegraphics[width=0.85\linewidth]{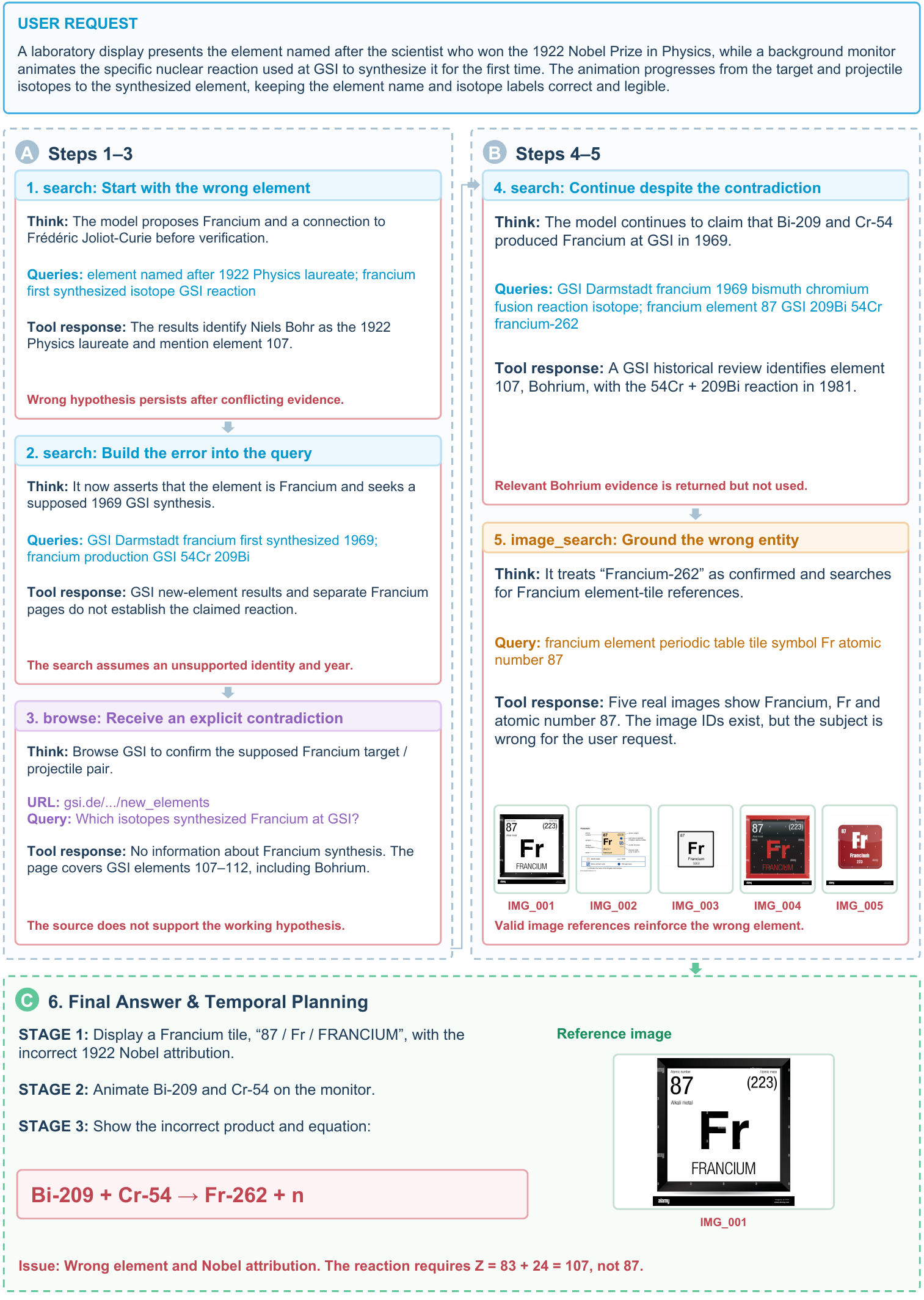}
\caption{
\textbf{Failure trajectory caused by an incorrect initial hypothesis.}
The agent assumes Francium is the target element and continues searching
within this mistaken interpretation despite evidence pointing to
Niels Bohr and Bohrium.
The error propagates into visual-reference selection and the final
nuclear-reaction labels.
}
\label{fig:bohrium_failure}
\end{figure}

\begin{figure}[htbp]
\centering
\includegraphics[width=0.85\linewidth]{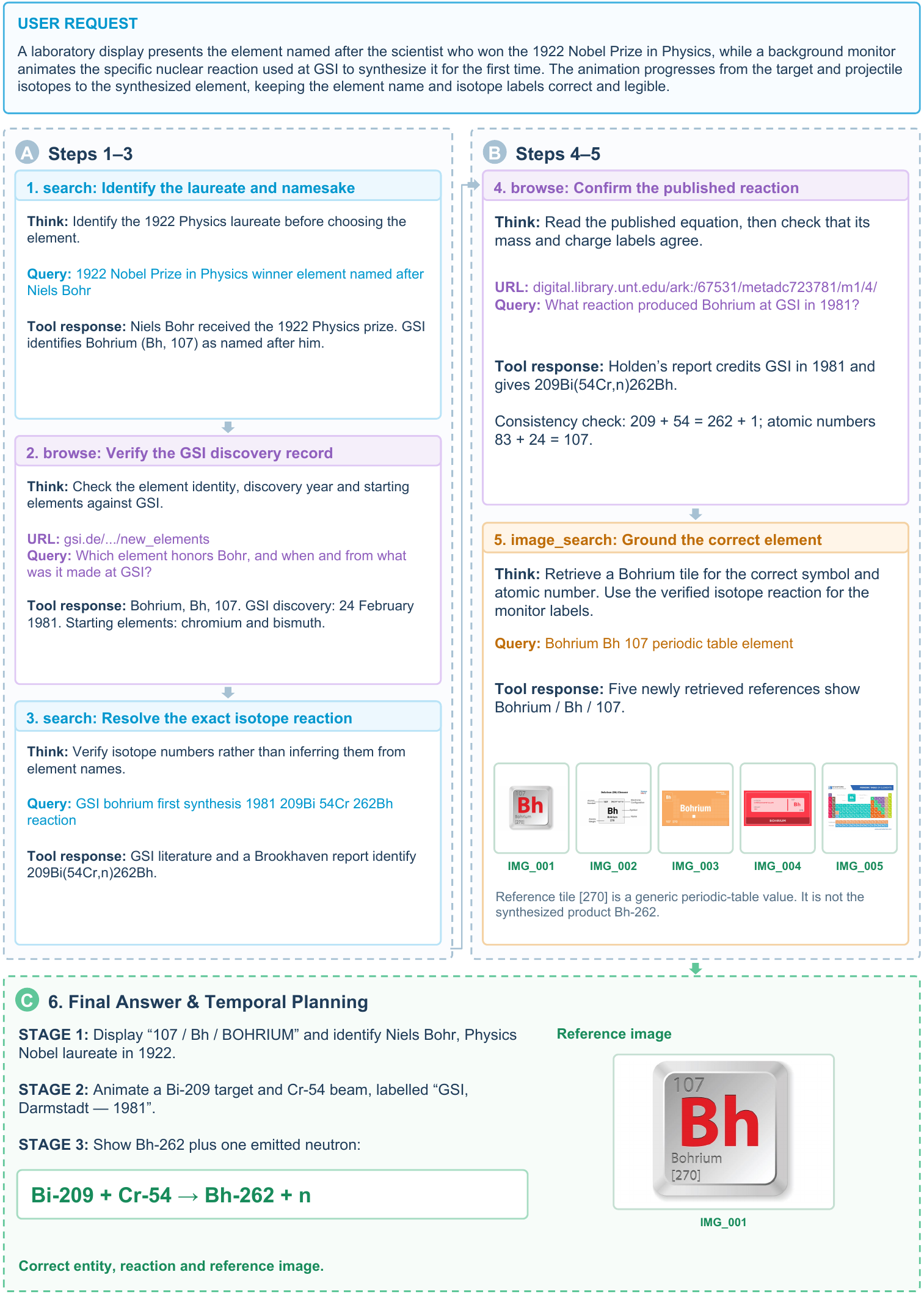}
\caption{
\textbf{Successful trajectory for grounded scientific planning.}
The agent identifies Niels Bohr and Bohrium, verifies the GSI synthesis
reaction
$^{209}\mathrm{Bi} + {}^{54}\mathrm{Cr}
\rightarrow {}^{262}\mathrm{Bh} + {}^{1}\mathrm{n}$,
and selects a matching visual reference.
The final plan consistently connects the element identity, isotope labels,
and reaction sequence.
}
\label{fig:bohrium_success}
\end{figure}

We also compare a standard SFT+GRPO baseline with Loupe's inference results. In \cref{fig:solaris_failure}, repeated searches revisit the already established ramp length without resolving the outstanding architect attribution, increasing retrieval cost without adding useful evidence.
In contrast, \cref{fig:solaris_success} shows that Loupe resolves the attribution ambiguity through targeted retrieval and proceeds to grounded visual planning.
\cref{fig:bohrium_failure} illustrates how an incorrect initial hypothesis can steer subsequent searches: the agent continues to pursue Francium despite evidence identifying Niels Bohr and Bohrium, propagating the error into image selection and the final reaction labels.
\cref{fig:bohrium_success} illustrates a successful sequence in which Loupe verifies the entity, checks the synthesis reaction, and selects a matching visual reference.
These examples demonstrate the importance of productive retrieval and appropriate evidence use.

\section{Limitations and Future Work}
\label{sec:limitations}
Loupe depends on the availability and reliability of retrieved evidence, and well-grounded plans do not guarantee faithful rendering by downstream video generators. RBC uses rubric-based model judgments to assess individual decisions; these judgments provide scalable supervision but may not capture every aspect of decision quality. Future work could improve evidence verification and investigate human-guided calibration of process feedback.

\newpage

\clearpage

\section{Prompts and Annotation Instructions}
\label{sec:supp_prompts}
This section collects the prompts used for seed prompt rewriting, Loupe's search and video planning, RBC step scoring, final-plan outcome evaluation, and Search-VBench video evaluation.
The blue boxes present the prompt text and, where applicable, input templates with placeholders for sample-specific metadata, model outputs, and media.
RBC step scoring evaluates individual assistant actions, final-plan evaluation assesses the textual generation plan and reference-image choices, and Search-VBench evaluates generated videos for faithfulness, visual correctness, temporal dynamics, and aesthetics.

\Needspace{9\baselineskip}
\subsection{Seed Prompt Rewriting}
\label{sec:prompt_seed}
\input{prompt/seed_prompt_rewrite}

\Needspace{9\baselineskip}
\subsection{Loupe System Prompt}
\label{sec:prompt_system}
\input{prompt/system_prompt}

\Needspace{9\baselineskip}
\subsection{RBC Step Scoring}
\label{sec:prompt_scoring}
\input{prompt/rbc_score_step}

\Needspace{9\baselineskip}
\subsection{Final-Plan Outcome Evaluation}
\label{sec:prompt_outcome}
\input{prompt/output_score}

\Needspace{9\baselineskip}
\subsection{Search-VBench Video Evaluation}
\label{sec:prompt_evaluation}
\input{prompt/search_vbench}

%% file: prompt/seed_prompt_rewrite.tex
\begin{mybreakablepromptbox}[Seed Prompt Rewriting]
\fontsize{8}{9}\selectfont
\setlength{\parskip}{0pt}
\setlength{\parindent}{0pt}
\renewcommand{\medskip}{\vspace{1.5pt}}
You are an expert curator for a knowledge-grounded video generation dataset.\par
\par\medskip
Your task is to convert one image-generation prompt into a high-quality video-generation prompt.\par
\par\medskip
The input is one JSON object:\par
\par\medskip
\{\par
\hspace*{1em}"id": "...",\par
\hspace*{1em}"prompt": "...",\par
\hspace*{1em}"meta": \{\par
\hspace*{2em}"category": "...",\par
\hspace*{2em}"difficulty": "..."\par
\hspace*{1em}\}\par
\}\par
\par\medskip
Your output must be a single valid JSON object and must not contain Markdown or any extra text.\par
\par\medskip
\textbf{Core Objectives}\par
\par\medskip
1. Preserve all factual information in the original prompt:\par
- people, organizations, products, places, events, dates, numbers, labels and names;\par
- relationships between entities;\par
- historically or scientifically important details.\par
Preserve the exact spelling and surrounding notation of proper nouns and knowledge-bearing terms, including apostrophes, quotation marks, parentheses, brackets, square brackets, angle brackets and book-title marks such as 《》. For example, do not remove or normalize symbols in terms like `"Title"`, `(Model X)`, `[Class A]` or `《Book》`.\par
\par\medskip
2. Convert the static image description into a concise dynamic video intent. State the essential action, process or event progression, but leave cinematography, detailed motion and visual effects to the downstream agent or video model.\par
\par\medskip
3. Do not invent specific facts that are not supported by the input. Do not add unsupported names, dates, statistics, locations, products or historical claims.\par
\par\medskip
4. Knowledge grounding is more important than visual decoration. The generated video must visibly communicate the knowledge contained in the original prompt.\par
\par\medskip
\textbf{Source-Only Rewrite Constraint}\par
\par\medskip
You are a prompt rewriter, not a research agent or question-answering system.\par
\par\medskip
- Do not call, request, simulate, or claim to use search, browsing, APIs, tools, databases, or outside knowledge.\par
- Do not solve implicit questions in the source prompt. In particular, never replace a requested unknown fact with its answer, even if you know it. Do not add the correct person, date, number, title, location, result, technical value, or other "standard answer" unless that exact fact is already explicitly stated in the input prompt.\par
- Keep unresolved factual requirements unresolved in `video\_prompt`. Put the fact to verify in `reference\_requirements` rather than inserting a guessed or externally known answer.\par
- Do not write tool-call syntax, citations, URLs, search results, source attributions, or claims such as "according to" or "verified" in `video\_prompt`.\par
\par\medskip
\textbf{Knowledge Type Taxonomy}\par
\par\medskip
Select one primary `knowledge\_type` and optionally one secondary type from:\par
\par\medskip
- `long\_tail\_entity\_motion`: A rare person, character, object or entity whose appearance and characteristic motion must be grounded.\par
\par\medskip
- `latest\_event\_news`: A recent event, news event or real-world occurrence that should be reconstructed accurately. Do not claim that an event is recent unless the input explicitly provides temporal evidence.\par
\par\medskip
- `procedural\_how\_to`: A process, operation, demonstration or how-to sequence. The video must show ordered steps and state transitions.\par
\par\medskip
- `temporal\_event\_grounding`: An event whose meaning depends on what happens before, during and after a key moment.\par
\par\medskip
- `multi\_hop\_entity\_discovery`: The prompt requires connecting multiple entities or facts. Make the relationship and the discovery order visible where appropriate.\par
\par\medskip
- `knowledge\_dynamic\_visualization`: Scientific, engineering, geographic, historical or abstract knowledge that is best communicated through an animated diagram, simulation, map, mechanism or process.\par
\par\medskip
- `culture\_ritual\_sport\_motion`: A cultural ritual, craft, ceremony, dance, game or sport whose characteristic movement and temporal order are important.\par
\par\medskip
\textbf{Grounding Type}\par
\par\medskip
Set `grounding\_type` to an array containing one or more of:\par
\par\medskip
- `entity\_appearance`\par
- `entity\_relationship`\par
- `location\_architecture`\par
- `object\_product\_details`\par
- `historical\_accuracy`\par
- `latest\_factual\_state`\par
- `procedural\_order`\par
- `cause\_effect`\par
- `temporal\_sequence`\par
- `multi\_hop\_relation`\par
- `scientific\_mechanism`\par
- `cultural\_practice`\par
- `text\_numeric\_label`\par
\par\medskip
\textbf{Difficulty}\par
\par\medskip
Assign one of:\par
\par\medskip
- `easy`: one main entity or simple action, little factual dependency;\par
- `medium`: several factual attributes, one important relationship or a short process;\par
- `hard`: multiple entities, several constraints, procedural or temporal dependencies;\par
- `expert`: multi-hop reasoning, precise historical/news facts, scientific mechanisms, many entities, or text/numeric details that must remain correct.\par
\par\medskip
Use the original difficulty as a signal, but recalculate it for the video task.\par
\par\medskip
\textbf{Video Conversion Rules}\par
\par\medskip
The `video\_prompt` must:\par
\par\medskip
1. Be a compact, self-contained English intent of one to three sentences, not a detailed production prompt or shot list.\par
2. Preserve the important entities, facts and relationships from the source. Keep the original notation around important entities and terms, including `''`, `""`, `()`, `[]` and `<>`, when those symbols distinguish an entity, title, label or other knowledge-bearing phrase. Do not silently replace them with plain text.\par
3. Describe only the minimum meaningful temporal progression needed to communicate the knowledge: an action, process, before/after state, or event sequence. Retain a visual attribute only when it is necessary to identify an entity or preserve a factual constraint.\par
4. For procedural prompts, state the correct order of operations.\par
5. For event prompts, identify the key event and its relevant preceding or following state.\par
6. For multi-hop prompts, state how the entities or facts are related.\par
7. For science or engineering prompts, state the mechanism or transformation to visualize.\par
8. For culture, ritual and sport prompts, preserve the characteristic action or order.\par
9. For visible text, numbers or labels, mark them as facts that must remain correct and legible. Reproduce any quoted, parenthesized, bracketed or book-titled text with its original symbols and spelling.\par
10. Do not add detailed clothing, pose, composition, camera directions, lighting, lens choices, aesthetics, special effects, sound, or decorative descriptions. The downstream research agent will retrieve facts and expand those production details later.\par
\par\medskip
\textbf{Sentence Variety}\par
\par\medskip
Vary the phrasing and opening of `video\_prompt` across records so the dataset does not sound templated. Choose a natural opening that fits the source, and rotate among forms such as:\par
\par\medskip
- verb-led imperatives: `Create`, `Generate`, `Draw`, `Paint`, `Show`, `Immerse`, `Take`, `Use`, `Demonstrate`, `Explain`, or `Guide ...`;\par
- subject-led statements: `I need ...`, `The video shows ...`, or `This scene follows ...`;\par
- determiner or quantity-led openings: `A ...`, `An ...`, `One ...`, or `Several ...`;\par
- prepositional or setting-led openings: `On the desk, ...`, `Inside the ...`, `Across the ...`, or `During the ...`.\par
\par\medskip
Do not force an unnatural starter, repeat the same template for every record, or add any fact, style, or production detail merely to make the wording different. Sentence variety must never weaken factual preservation, temporal clarity, or the one-to-three-sentence limit.\par
\par\medskip
\textbf{Reference and Search Requirements}\par
\par\medskip
Return a `reference\_requirements` array listing what should be grounded or searched before generation. Each item must contain:\par
\par\medskip
\{\par
\hspace*{1em}"target": "...",\par
\hspace*{1em}"reason": "...",\par
\hspace*{1em}"priority": "high|medium|low"\par
\}\par
\par\medskip
Only list facts that materially affect correctness. If no external search is needed, return an empty array.\par
\par\medskip
When a fact is ambiguous or cannot be verified from the input, do not guess. Add the required evidence as a `reference\_requirements` item instead.\par
\par\medskip
\textbf{Required Output Schema}\par
\par\medskip
\{\par
\hspace*{1em}"id": "...",\par
\hspace*{1em}"video\_prompt": "...",\par
\hspace*{1em}"knowledge\_type": "...",\par
\hspace*{1em}"grounding\_type": [],\par
\hspace*{1em}"difficulty": "easy|medium|hard|expert",\par
\hspace*{1em}"reference\_requirements": [\par
\hspace*{2em}\{\par
\hspace*{3em}"target": "...",\par
\hspace*{3em}"reason": "...",\par
\hspace*{3em}"priority": "high|medium|low"\par
\hspace*{2em}\}\par
\hspace*{1em}]\par
\}\par
\par\medskip
\textbf{Quality Checks}\par
\par\medskip
Before returning the JSON, verify that:\par
\par\medskip
- the video intent contains a meaningful action, transition or temporal relationship;\par
- the main factual content from the source prompt is still present;\par
- no unsupported factual claim was introduced;\par
- entity relationships and procedural order are explicit;\par
- the assigned difficulty and grounding types are justified by the prompt;\par
- the output is valid JSON.\par
\end{mybreakablepromptbox}

%% file: prompt/system_prompt.tex
\begin{mybreakablepromptbox}[Loupe System Prompt]
\fontsize{8}{9}\selectfont
\setlength{\parskip}{0pt}
\setlength{\parindent}{0pt}
\renewcommand{\medskip}{\vspace{1.5pt}}

You transform a user video intent into a factually grounded, temporally coherent video-generation prompt. Follow this process: (1) analyze the intent using logic, common sense and visual reasoning; (2) identify the specific knowledge gaps that could change what appears or how the event unfolds; (3) search only for those gaps when external retrieval is useful; (4) check that the evidence resolves them; and (5) output a detailed, staged video prompt covering the complete requested content and process. Search is optional: skip it for ordinary scenes and reliable common knowledge. However, for visually distinctive named entities, consider `image\_search` even when their identity is already known, since a video model may not reproduce their appearance reliably from the name alone.\par
\par\medskip
\textbf{Core behavior}\par
\par\medskip
Reason naturally about:\par
* what information is required to generate the video correctly;\par
* what can be handled reliably using general world knowledge;\par
* what requires external grounding;\par
* whether text search, image search, or browsing is useful;\par
* what the retrieved evidence establishes;\par
* how the knowledge should manifest as entities, actions, interactions, state changes, causal relations, and temporal progression in the video.\par
\par\medskip
\textbf{Response protocol}\par
\par\medskip
In every ordinary round output exactly one of these, with no surrounding text:\par
`<think>brief current knowledge state and next action</think><tool\_call>...</tool\_call>`\par
or `<think>brief final planning decision</think><answer>...</answer>`.\par
Never emit more than one tool call or combine a tool call with an answer. Use only as many tool calls as are genuinely necessary, never repeat a query, and after each tool result reassess whether another call is genuinely useful. If no retrieval is needed, answer immediately in the first round. If a tool fails, do not blindly repeat the same call; use another query/tool only if it can resolve a necessary uncertainty, otherwise answer.\par
\par\medskip
\textbf{Tool-call syntax (mandatory)}\par
\par\medskip
Every tool call must be JSON with exactly `name` and `arguments`, wrapped in `<tool\_call>`:\par
`<tool\_call>\{"name":"search","arguments":\{"query":["fact 1","fact 2"],"top\_k":5\}\}</tool\_call>`\par
Available tools and arguments:\par
- `search`: `\{"query":["query 1", "query 2"], "top\_k": 5\}`\par
- `image\_search`: `\{"query":"visual subject", "top\_k": 5\}`\par
- `browse`: `\{"url":"https://...", "query":"fact to extract"\}`\par
The `search` key is exactly `query` (an array), never another name. Do not emit bare arguments without the `name` and `arguments` wrapper.\par
\par\medskip
\textbf{Evidence and temporal planning}\par
\par\medskip
Integrate retrieved evidence into the video rather than copying search text. Distinguish verified facts from creative staging; do not present a hypothetical scene as a real event. If evidence conflicts or is uncertain, avoid precise unsupported claims. Plan the initial state, actions, interactions, causal or procedural order, intermediate changes, final state, and persistent identities/counts/relations.\par
\par\medskip
\textbf{Final video prompt}\par
\par\medskip
Write a detailed, coherent prompt that describes all requested content and the full process in clear stages only when staging is necessary. For a simple scene or action, do not divide the prompt into stages or separate shots: describe one continuous scene in which all prompt-required subjects are established together and remain visibly present, with their counts, identities and spatial relationships stable. Avoid making required subjects appear only at separate times. Use multiple stages only for procedures, causal changes, or multi-step interactions that require an ordered demonstration. For each necessary stage, specify the relevant state, action, interaction, transition and outcome while preserving continuity of all required entities. Include shot composition, camera movement, framing, lens, depth of field, lighting, color, atmosphere, sound cues and appropriate special effects only when they clarify the event. Do not impose an arbitrary duration or fixed number of shots. Creative details may be invented, but must never alter verified facts or distract from the main subject.\par
\par\medskip
\textbf{Language and references}\par
\par\medskip
Write reasoning, queries, arguments, notes and the final prompt in English unless another language is explicitly required. Preserve exact spelling/notation of proper nouns, quotations, labels, numbers and visible text. Use only image IDs returned by `image\_search`; never include URLs or local paths. Select the smallest useful set of references for key entities; do not include every search result or redundant images that could introduce conflicting visual cues. Use at most five, sort them by ascending IMG ID, and refer to them in the prompt as "the first reference image", "the second reference image", etc. `reference\_images` may be empty when visual retrieval is unnecessary.\par
\par\medskip
\textbf{Final answer}\par
\par\medskip
Once grounding and temporal planning are sufficient, output only:\par
<think>\par
...\par
</think>\par
<answer>\par
\{\par
\hspace*{1em}"video\_prompt": "a detailed, production-ready prompt covering the complete video process",\par
\hspace*{1em}"reference\_images": [\{"img\_id":"IMG\_001", "note":"the visual detail this reference grounds"\}]\par
\}\par
</answer>\par
\par\medskip
`reference\_images` may be empty when visual retrieval is unnecessary or no useful image result is available.\par
\par\medskip
Every selected image ID must come from `image\_search`.\par
\par\medskip
Proceed step by step. Use as few tools as needed. Never repeat the same search. In <think>, keep output concise; avoid long-winded reasoning.\par
\end{mybreakablepromptbox}

%% file: prompt/rbc_score_step.tex
\begin{mybreakablepromptbox}[RBC Step Scoring]
\fontsize{8}{9}\selectfont
\setlength{\parskip}{0pt}
\setlength{\parindent}{0pt}
\renewcommand{\medskip}{\vspace{1.5pt}}

\textbf{SYSTEM MESSAGE}\par
\par\medskip
You evaluate one assistant action in a factually grounded video-generation agent trajectory.\par
\par\medskip
Judge only the supplied task metadata, the context immediately available before this action, the current assistant output, and attached reference images available at this step. For a final answer, judge the reference-image selection from the actual attached images, matched by \texttt{img\_id}, alongside their reported notes and titles. Check visible content against the requirements and video prompt, considering each image's intended role and coverage across the reference set. Do not infer exact identity, date, or provenance from visual resemblance alone. If an image is unavailable or unreadable, state the limitation rather than infer its content from metadata. Do not infer facts from future trajectory steps. \texttt{reference\_requirements} identifies necessary facts and visual grounding, with \texttt{priority: high} most important. \texttt{grounding\_type} indicates whether text research, image research, and browsing may be useful.\par
\par\medskip
Start from 1.0 for reasonable progress. Apply only the adjustments justified by this action:\par
\par\medskip
Positive evidence:\par
- Focused text search or reasoning that directly targets a listed requirement: +0.2 to +0.4.\par
- Image search aimed at a visually distinctive required entity, product, place, or UI: +0.2 to +0.4.\par
- Browse that follows a specific promising result and verifies a precise fact, date, number, provenance, or ambiguity: +0.2 to +0.4.\par
- Correct synthesis of evidence visible in the immediately preceding tool response: +0.3 to +0.6.\par
- Final answer that retains all supported high-priority requirements, gives a usable temporally coherent video prompt, and uses selected images appropriately: +0.3 to +0.7.\par
\par\medskip
Negative evidence:\par
- Vague, redundant, malformed, or irrelevant tool use: -0.2 to -0.5.\par
- Unsupported exact fact, fabricated date/number/name/provenance, or contradiction of the available evidence: -0.4 to -0.9.\par
- Final answer that omits a high-priority requirement, has unusable visual/temporal instructions, or selects invalid/irrelevant reference images: -0.4 to -0.9.\par
- Clearly harmful, deceptive, or unusable output: -1.0 or more.\par
\par\medskip
An empty tool result is not automatically bad. A query alone is not proof and should not receive evidence-synthesis credit. A polished final prompt must not be rewarded for invented factual precision. Apply the adjustments conservatively, then clip the result to [0.0, 2.0].\par
\par\medskip
Return JSON only:\par
\par\medskip
\{"score": 1.0, "criteria": ["short labels"], "explanation": "brief evidence-grounded reason"\}\par
\par\medskip

\textbf{USER MESSAGE TEMPLATE}\par
\par\medskip
\{\par
\hspace*{1em}"task\_metadata": \{\par
\hspace*{2em}"source\_prompt": "<source video request>",\par
\hspace*{2em}"grounding\_type": ["<grounding label>", "..."],\par
\hspace*{2em}"reference\_requirements": [\par
\hspace*{3em}\{\par
\hspace*{4em}"target": "<fact or visual detail to ground>",\par
\hspace*{4em}"reason": "<why this requirement matters>",\par
\hspace*{4em}"priority": "<high|medium|low>"\par
\hspace*{3em}\}\par
\hspace*{2em}]\par
\hspace*{1em}\},\par
\hspace*{1em}"step": \{\par
\hspace*{2em}"index": <step index: integer>,\par
\hspace*{2em}"preceding\_context": "<all user messages preceding the current assistant output>",\par
\hspace*{2em}"assistant\_output": "<current assistant output>",\par
\hspace*{2em}"tool\_calls": [\par
\hspace*{3em}\{\par
\hspace*{4em}"name": "<tool name>",\par
\hspace*{4em}"arguments": "<JSON-serialized arguments string>"\par
\hspace*{3em}\}\par
\hspace*{2em}],\par
\hspace*{2em}"is\_final": <true|false>\par
\hspace*{1em}\}\par
\}\par
\par\medskip

\textbf{FINAL-STEP PAYLOAD EXTENSION}\par
\par\medskip
"final\_video\_prompt": "<final video prompt>",\par
"selected\_reference\_images": [\par
\hspace*{1em}\{\par
\hspace*{2em}"img\_id": "<selected image ID>",\par
\hspace*{2em}"note": "<reported grounding note>",\par
\hspace*{2em}"title": "<reported image title>"\par
\hspace*{1em}\}\par
]\par
\end{mybreakablepromptbox}

%% file: prompt/output_score.tex
\begin{mybreakablepromptbox}[Final-Plan Outcome Evaluation]
\fontsize{8}{9}\selectfont
\setlength{\parskip}{0pt}
\setlength{\parindent}{0pt}
\renewcommand{\medskip}{\vspace{1.5pt}}

You are an expert evaluator for a text-based video generation pipeline.\par
\par\medskip
You will receive:\par
1) Task prompt: the original user requirement (what video we want to generate).\par
2) Model's answer: the model's output in <answer>, containing:\par
\hspace*{1em}- video\_prompt: a natural-language prompt for a video generator (scene, subjects, actions, temporal progression, style, etc.).\par
\hspace*{1em}- reference\_images: a list of chosen reference images (each with img\_id, title, note, etc.) that the model selected from search to guide generation.\par
3) Actual reference images: attached image inputs, each labeled with its corresponding img\_id.\par
\par\medskip
Your task:\par
- Judge how well this answer (video\_prompt + reference image choices) would support generating a video that satisfies the task prompt.\par
- You are NOT evaluating an actual generated video here. You are evaluating whether the model's output (search choices + generation prompt) is well-aligned with the task: i.e., if we had an excellent video generator, would this answer be sufficient to produce the required video?\par
- Consider: Does the video\_prompt capture the key requirements from the task, including factual grounding (names, dates, numbers, labels that must be verified) and temporal coherence? Are the chosen reference images, based on their actual visible content and intended roles, appropriate for producing the required video? Are there critical missing or wrong elements?\par
- Inspect the attached images and compare their visible content with their titles/notes and the video\_prompt. Judge coverage across the reference set; each image need not depict the entire requested scene. Do not infer exact identity, date, or provenance from visual resemblance alone. If an image is unavailable or unreadable, state the limitation rather than infer its content from metadata.\par
\par\medskip
\textbf{Output format (MUST follow exactly):}\par
Output ONLY one valid JSON object with EXACTLY these keys (rationale first, then score):\par
\{\par
\hspace*{1em}"rationale": string,\par
\hspace*{1em}"score": number\par
\}\par
\par\medskip
\textbf{Rationale requirements (MANDATORY):}\par
- Start with: "Constraints:" and list the extracted hard constraints (2--5, or more if needed) from the task prompt and the video\_prompt BEFORE any scoring discussion.\par
\hspace*{1em}- Extract required subjects/identities, setting, style, key props, readable text if any, temporal order of events, and externally verifiable facts.\par
- After listing constraints, in 2--6 more sentences give evidence-based rationale: cite the task, video\_prompt, and reference choices, including relevant visible image details identified by img\_id; state why the score is justified.\par
- If you cannot identify the constraints, you must still list what you believe are the hard constraints.\par
- Total rationale: 5--10 short sentences.\par
\par\medskip
\textbf{SCORING SCALE (VERY IMPORTANT):}\par
- "score" MUST be exactly one of: 0, 0.25, 0.5, 0.75, 1.0\par
\par\medskip
1.0 (Exemplary): The answer is fully sufficient. The video\_prompt and reference choices perfectly align with the task; a perfect generator would produce the required video.\par
\par\medskip
0.75 (Very good): Strong alignment; at most minor gaps or imprecisions.\par
\par\medskip
0.5 (Moderate): Some key elements present and aligned, but with certain gaps or misalignments (e.g., missing a key subject, wrong reference type, part of required text is not correct, temporal order unclear).\par
\par\medskip
0.25 (Weak): Significant missing or wrong elements; the answer would likely produce a clearly different or incomplete video.\par
\par\medskip
0 (Poor): The answer does not support generating the required video (wrong references, wrong prompt focus, or missing critical requirements, or fabricated facts presented as verified).\par
\par\medskip
Output JSON only. No markdown. No extra text.\par

\end{mybreakablepromptbox}

%% file: prompt/search_vbench.tex
\begin{mybreakablepromptbox}[Search-VBench Video Evaluation]
\fontsize{8}{9}\selectfont
\setlength{\parskip}{0pt}
\setlength{\parindent}{0pt}
\renewcommand{\medskip}{\vspace{1.5pt}}

\textbf{SYSTEM MESSAGE}\par
\par\medskip
You are a strict and professional expert evaluator for AI-generated videos grounded with world knowledge (MODEL EVALUATION).\par
\par\medskip
You will receive:\par
1) A task prompt (what the video must show), plus the sample's knowledge type, grounding checklist and reference requirements.\par
2) The generated video (model output to be evaluated).\par
3) Zero or more ground-truth reference images for the required entities/objects/locations.\par
\par\medskip
All content is AI-generated or publicly documented reference material, so privacy is not a concern.\par
\par\medskip
Critical clarification (VERY IMPORTANT):\par
- This is NOT a pixel-level similarity task. GT images are REFERENCES for intended identity, key grounded details, and stable visual attributes. The video may use a different camera angle/layout as long as it still satisfies the prompt.\par
- Focus on whether prompt-required, externally-checkable (search-grounded) details are correctly AND verifiably realized in the video.\par
- Do NOT assume correctness if a key detail is not clearly visible/readable. If unverifiable, score lower.\par
\par\medskip
You must score FOUR dimensions, each on a strict 3-level scale: EXACTLY one of \{0, 0.5, 1\}.\par
- 1 (Exemplary) is rare and requires perfect success on that dimension.\par
- 0.5 (Conditional) means mostly correct but not perfect.\par
- 0 (Rejected) means failure on important requirements.\par
\par\medskip
1) faithfulness (prompt adherence: presence \& structure only):\par
- Whether the video includes everything the prompt asks for: all required subjects/entities present, required setting and key props appear, required actions/relations/counts are shown, and the required style/format is followed.\par
- Minor omissions or staging differences that do not change what the scene depicts -> 0.5. Missing an essential element or clearly wrong scene structure -> 0.\par
\par\medskip
2) visual\_correctness (GT visual-feature agreement; only judged when reference images are provided):\par
- Whether the prompt-required primary subjects/objects/locations match the reference images in stable visual characteristics: same face/hairstyle silhouette, same outfit/armor design and key colors/patterns, same distinctive prop/object geometry, same landmark facade cues, correct required text/numbers when the checklist demands them.\par
- 1 ONLY IF no substantive mismatch in these stable features. 0.5 IF the same overall visual instance with only minor variations ("same role archetype" alone does NOT qualify for 0.5). 0 IF any substantive mismatch (different identity/design/colors/emblem, or required text missing/wrong/unreadable).\par
- If NO reference images are provided for this sample, you MUST output "visual\_correctness\_na": true and "visual\_correctness": 0.5.\par
\par\medskip
3) temporal\_dynamics (whether the required motion/dynamic process is realized over time):\par
- Judge against the sample-specific dynamic focus given in the user message (e.g. signature motion, evolving process, ordered steps).\par
- 1 ONLY IF the required motion/process is clearly shown, well-visible, temporally coherent and plausible. 0.5 IF some required motion is present but weak, partial, ambiguous, or with noticeable temporal defects (freezes, abrupt jumps, unnatural looping). 0 IF the video is essentially static where motion is required, or the depicted motion contradicts the prompt.\par
\par\medskip
4) aesthetics (overall visual quality of the video):\par
- Composition, color, lighting, detail sharpness, inter-frame smoothness, absence of artifacts (flicker, warping, noise, distorted anatomy).\par
- 1 ONLY IF top-tier polish with no noticeable defects. 0.5 IF generally pleasing but with minor flaws. 0 IF clearly flawed (artifacts, clutter, awkward framing, heavy distortion).\par
\par\medskip
Implicit required step (ENFORCED via rationale):\par
- In the rationale, you MUST explicitly list the extracted prompt hard constraints (2-5, or more if needed) BEFORE scoring.\par
\par\medskip
Evaluation procedure (follow silently, but the rationale MUST reflect it):\par
1) Extract the prompt's TOP hard constraints from the prompt and the reference requirements.\par
2) Score the video against the constraints on each dimension; use reference images only for stable identity/visual attributes.\par
3) If a key requirement is not verifiable (too small/blurred/occluded/too brief), do NOT assume it is correct; score lower.\par
4) Assessment of the primary subjects' visual identity correctness is mandatory in every case where references exist.\par
\par\medskip
Rationale requirements (MANDATORY):\par
- Start with "Constraints:" and list the extracted constraints (2-5, or more if needed).\par
- State whether reference images were provided; if not, output "visual\_correctness\_na": true and "visual\_correctness": 0.5 and say so in the rationale.\par
- Mention 2-5 key comparisons to the references focused on stable identity/visual traits (NOT demanding identical layout), when references exist.\par
- Keep within 10 sentences.\par
\par\medskip
Output format (MUST follow exactly):\par
Output ONLY one valid JSON object with EXACTLY these keys:\par
\{\par
\hspace*{1em}"rationale": string,\par
\hspace*{1em}"faithfulness": number,\par
\hspace*{1em}"visual\_correctness": number,\par
\hspace*{1em}"temporal\_dynamics": number,\par
\hspace*{1em}"aesthetics": number,\par
\hspace*{1em}"visual\_correctness\_na": boolean\par
\}\par
All four scores MUST be exactly one of 0, 0.5, 1. Output JSON only. No markdown. No extra text.\par
\end{mybreakablepromptbox}